\documentclass[10pt,journal]{IEEEtran}
\usepackage{amsmath,amssymb,amsfonts}
\usepackage{algorithm}
\usepackage{algorithmicx}
\usepackage{algpseudocode}
\usepackage{makecell}
\usepackage{wrapfig}
\usepackage{graphicx}
\allowdisplaybreaks[4]
\usepackage{subfig}
\usepackage{multirow}
\usepackage{colortbl}
\usepackage{textcomp}
\usepackage{xcolor}
\usepackage{pdfpages}
\usepackage{booktabs}
\usepackage{hyperref}
\usepackage{enumitem}
\usepackage{orcidlink}
\usepackage{cite}
\hypersetup{colorlinks=true,linkcolor=black,citecolor=black,urlcolor=black}

\newtheorem{assumption}{Assumption}
\newtheorem{proposition}{Proposition}
\newtheorem{lemma}{Lemma}
\newtheorem{remark}{Remark}

\newtheorem{theorem}{Theorem}

\newtheorem{definition}{Definition}
\newenvironment{proof}{{\it Proof}.\ }{\hfill $\blacksquare$\par}

\newcommand{\refappendix}[1]{\hyperref[#1]{Appendix \ref*{#1}}}

\def\BibTeX{{\rm B\kern-.05em{\sc i\kern-.025em b}\kern-.08em
    T\kern-.1667em\lower.7ex\hbox{E}\kern-.125emX}}

\begin{document}

\title{FedAgg: Adaptive Federated Learning with Aggregated Gradients}

\author{Wenhao~Yuan\orcidlink{0009-0001-6625-7496},~\IEEEmembership{Student Member, IEEE}, and Xuehe~Wang\orcidlink{0000-0002-6910-468X},~\IEEEmembership{Senior Member, IEEE}

\IEEEcompsocitemizethanks{\IEEEcompsocthanksitem Wenhao Yuan and Xuehe Wang are with the School of Artificial Intelligence, Sun Yat-sen University, Zhuhai 519082, China. Email Address: \href{mailto:yuanwh7@mail2.sysu.edu.cn}{yuanwh7@mail2.sysu.edu.cn}, \href{mailto:wangxuehe@mail.sysu.edu.cn}{wangxuehe@mail.sysu.edu.cn}.
}
}

\IEEEtitleabstractindextext{
\begin{abstract}
Federated Learning (FL) has emerged as a crucial distributed training paradigm, enabling discrete devices to collaboratively train a shared model under the coordination of a central server, while leveraging their locally stored private data. Nonetheless, the non-independent-and-identically-distributed (Non-IID) data generated on heterogeneous clients and the incessant information exchange among participants may significantly impede training efficacy, retard the model convergence rate and increase the risk of privacy leakage. To alleviate the divergence between the local and average model parameters and obtain a fast model convergence rate, we propose an adaptive \underline{FED}erated learning algorithm called FedAgg by refining the conventional stochastic gradient descent (SGD) methodology with an \underline{AG}gregated \underline{G}radient term at each local training epoch and adaptively adjusting the learning rate based on a penalty term that quantifies the local model deviation. To tackle the challenge of information exchange among clients during local training and design a decentralized adaptive learning rate for each client, we introduce two mean-field terms to approximate the average local parameters and gradients over time. Through rigorous theoretical analysis, we demonstrate the existence and convergence of the mean-field terms and provide a robust upper bound on the convergence of our proposed algorithm. The extensive experimental results on real-world datasets substantiate the superiority of our framework in comparison with existing state-of-the-art FL strategies for enhancing model performance and accelerating convergence rate under IID and Non-IID datasets.
\end{abstract}

}

\maketitle

\IEEEdisplaynontitleabstractindextext

\IEEEpeerreviewmaketitle

\section{Introduction} \label{introduction}

With the flourishing proliferation of edge devices, including but not limited to smartphones and wearable devices, etc, the deluge of private data originating from these geographically distributed nodes has swelled exponentially. Tremendous repositories of data provide more opportunities for artificial intelligence (AI) researchers, enabling them to delve deeper into the latent value and potential of data, thereby facilitating the training of sophisticated multi-modal models such as GPT-4. However, the transmission of voluminous datasets stored in edge devices to the centralized cloud is impractical and entails considerable expenditure of communication resources. Furthermore, the advent of stringent regulatory frameworks safeguarding the data privacy of mobile devices, exemplified by the General Data Protection Regulation (GDPR) \cite{regulation2016regulation}, constructs a formidable barrier for the realm of AI applications to access and leverage these private data. The dual challenges of privacy protection and big data have boosted the profound change of brand-new technologies that aggregate and model the data under the premise of meeting data privacy, security, and regulatory requirements \cite{sun2020adaptive}.

Federated learning serves as a promising paradigm to tackle distributed machine learning tasks and achieves multi-fold performance benefits including personal data privacy preservation and training efficiency improvement\cite{mcmahan2017communication}. Specifically, in the FL model training process, each participating client undertakes one or several epochs of mini-batch stochastic gradient descent (SGD) with their private dataset and subsequently uploads the model parameter instead of the raw data to the central server. After collecting and aggregating all local parameters, the central server updates the global parameter and distributes the refined global model to all participants for the next round of training. This above iterative process continues until the global model converges, reaching the specified maximum global iteration or attaining the required model accuracy threshold~\cite{zhang2021adaptive}. 

However, the conventional FL framework is confronted with salient challenges, with client heterogeneity being a pivotal one, encompassing a spectrum of disparities ranging from statistical heterogeneity to computational and communication heterogeneity. Pertaining to statistical heterogeneity, the data is independently generated on each client's geographically dispersed device and potentially yields substantial variations in data volume and label distribution, which may risk the emergence of client drifting thus precipitating a degradation in model performance and decelerating the convergence rate of the global model \cite{karimireddy2020scaffold, li2023anarchic}. Moreover, despite participants executing multiple epochs of local computation before sharing their localized models for aggregation, a significant communication overhead remains a recognized bottleneck in FL and the model training efficiency could suffer due to the limited resources such as bandwidth, energy, and power \cite{bonawitz2019towards}. 

\begin{figure*}[t]
\setlength{\abovecaptionskip}{2pt} 
    \centering
    \begin{minipage}{160pt}
    \centerline{\includegraphics[width=1.0\textwidth, trim=0 0 0 0,clip]{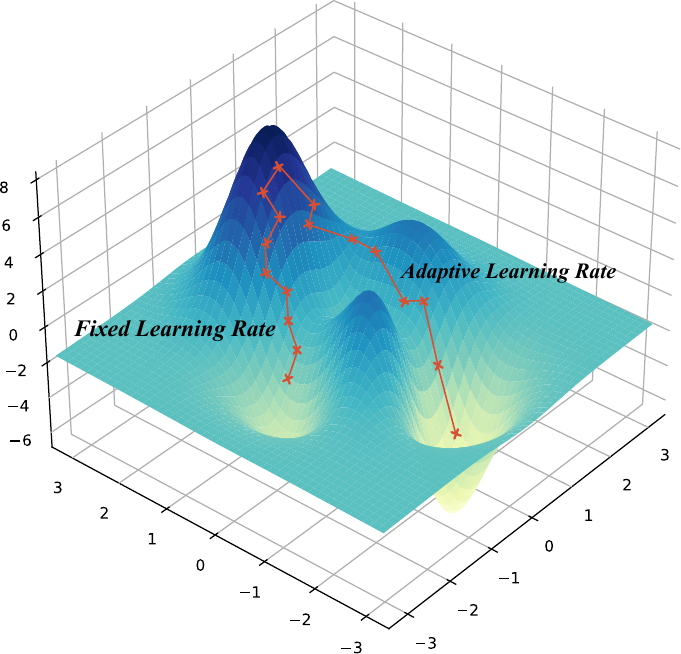}}
        \caption{The gradient descend comparison of the fixed and adaptive learning rate strategies.}
        \label{client_dev}
    \end{minipage}
    \hspace{5pt}
    \begin{minipage}{340pt}
        \subfloat{
            \label{MNIST_beta_acc}
            \includegraphics[width=0.5\textwidth, trim=30 0 70 20,clip]{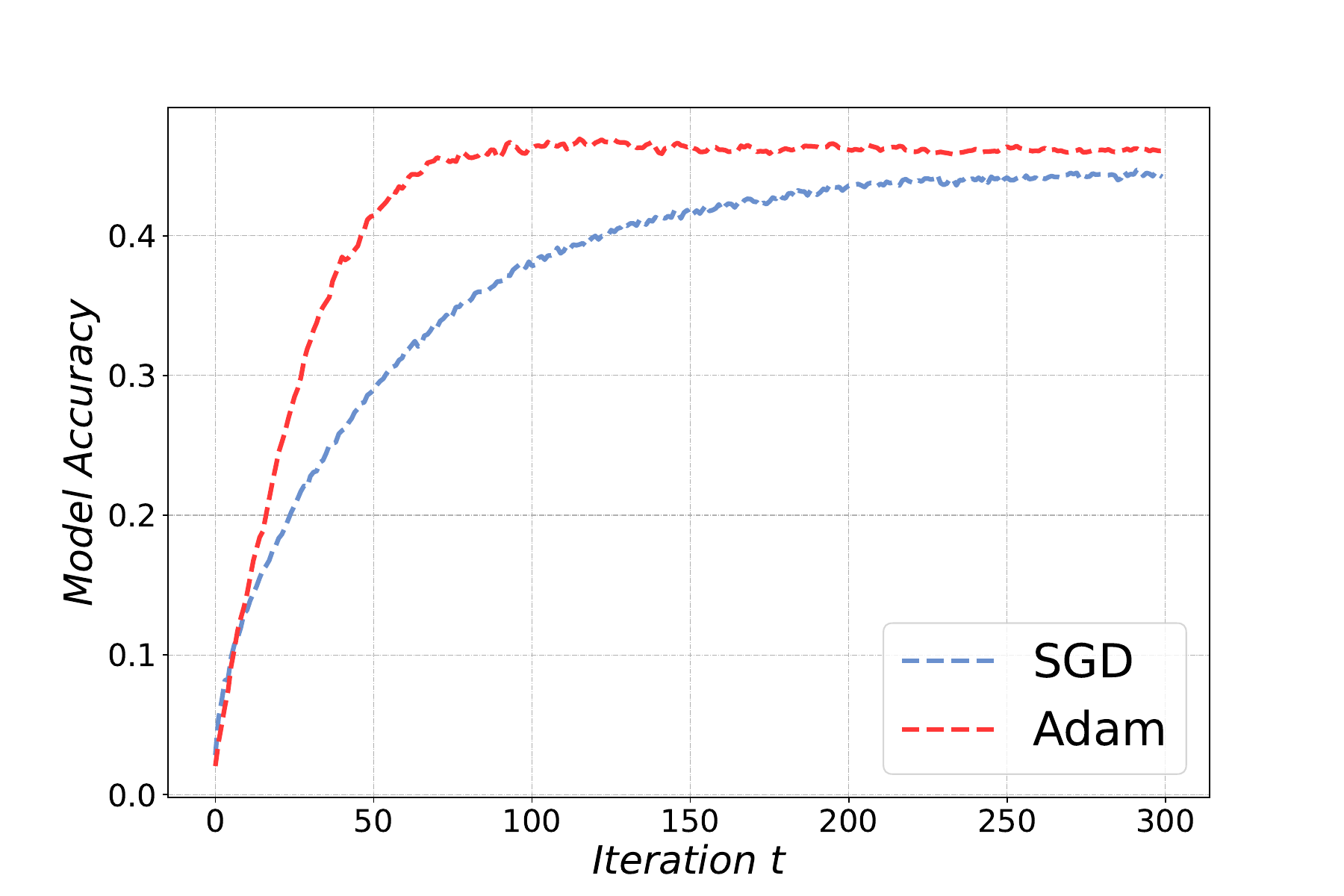}}
        \subfloat{
            \label{MNIST_beta_loss}
            \includegraphics[width=0.5\textwidth, trim=30 0 70 20,clip]{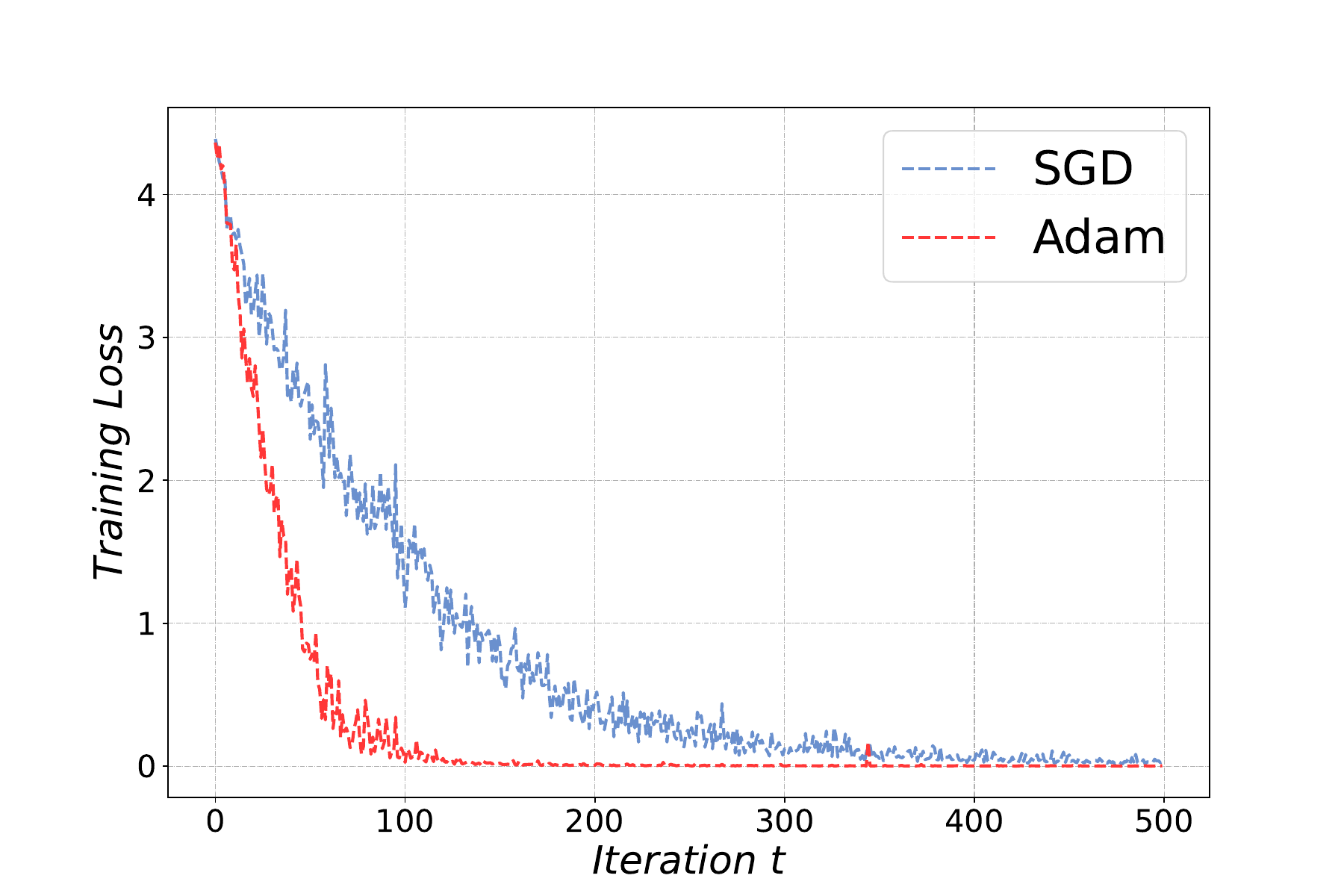}} 
         \caption{Numerical analysis of model accuracy and training loss curves on the CIFAR-100 dataset featuring IID data distribution. The results underscore the substantial impact of employing the adaptive learning rate scheme based on adaptive optimizer Adam, which enhances model performance and convergence rate.}
        \label{beta}
    \end{minipage}
\vspace{-15pt}
\end{figure*}

To address the aforementioned challenges, substantial research endeavors have been dedicated. The seminal work in \cite{mcmahan2017communication} introduces FedAvg, a foundational approach for federated optimization that highly depends on iterative model averaging and partial participation, thereby significantly decreasing the communication overhead. FedAvg demonstrates commendable efficiency for IID datasets, but its performance degrades and is unstable under the Non-IID settings. To mitigate the negative effects of client heterogeneity, researchers have proposed the asynchronous FL approach \cite{liu2024fedasmu, zhang2023timelyfl,hu2023scheduling} and adopted the client selection mechanism \cite{xu2020client,chai2020tifl,wang2020optimizing}. The asynchronous FL approach facilitates global aggregation upon the receipt of a single local update from any client. However, this approach risks the global model performance, which may be disproportionately influenced by a single client's local model, leading to suboptimal convergence. Furthermore, partial updates (i.e., client selection mechanism design) select a subset of clients to contribute to the global aggregation, which may introduce significant bias, as the global model predominantly reflects the characteristics of the selected clients, thus compromising model accuracy and raising fairness concerns. 

Nevertheless, in FL systems, the potential of adaptive learning rate-based algorithms in FL remains largely underexplored. Current literature often undervalues the pivotal role of the learning rate, a hyperparameter that requires meticulous tuning to accelerate the convergence speed and FL model performance. In Fig.~\ref{client_dev}, we observe that employing adaptive learning rates on the first-order optimization algorithms is instrumental in enhancing the efficiency of FL models, especially for the large-scale FL optimization tasks, as highlighted in \cite{zhou2018adashift}. Fig.~\ref{beta} shows a quantitative analysis of model accuracy and training loss curves on the CIFAR-100 dataset with IID data distribution and 20\% client participation ratio. The results underscore the profound influence of adaptive learning rate mechanisms, which not only significantly improve model performance but also demonstrate its efficacy in accelerating the convergence rate and enhancing training stability. 

In this paper, to achieve better FL model performance, we propose an adaptive learning rate scheme by considering the aggregated gradients of all clients in the local model updating process and the deviations between the local and average parameters. Our innovation points and main contributions are summarized as follows:
\begin{itemize} 
\item \emph{Adaptive learning rate design for fast model convergence:} To the best of our knowledge, this paper is the first to investigate the closed-form design of an adaptive learning rate for each client at each local epoch by introducing an aggregated gradient term into the SGD-based local updating rules. Furthermore, to mitigate the adverse effects of client drifting due to heterogeneity, we incorporate each client’s local parameter deviations into its long-term objective optimization function for adaptive learning rate design, thereby enhancing the performance of the FL model and accelerating the convergence rate. 

\item \emph{Decentralized adaptive learning rate design via mean-field terms:} To address the challenge of inter-client information exchange during local training epochs, we introduce two mean-field estimators to approximate the average local parameters and gradients of all clients over time, based on which, a decentralized adaptive learning rate is designed for each client. Through theoretical analysis, we demonstrate the existence and convergence of the mean-field terms and propose an iterative algorithm for their determination.

\item \emph{Convergence analysis and performance evaluation:} We conduct a rigorous theoretical analysis for the model convergence upper bound of our proposed FedAgg with the aggregated gradients scheme. Extensive experiments conducted on various real-world datasets (MNIST, EMNIST-L, CIFAR-10, and CIFAR-100) under both IID and Non-IID settings demonstrate the superiority of our proposed FedAgg algorithm over state-of-the-art FL benchmarks, by achieving a faster convergence rate and improving model accuracy.
\end{itemize}

\begin{figure*}[t]
\setlength{\abovecaptionskip}{0pt}
\centerline{\includegraphics[width=0.95\textwidth, trim=75 65 70 60,clip]{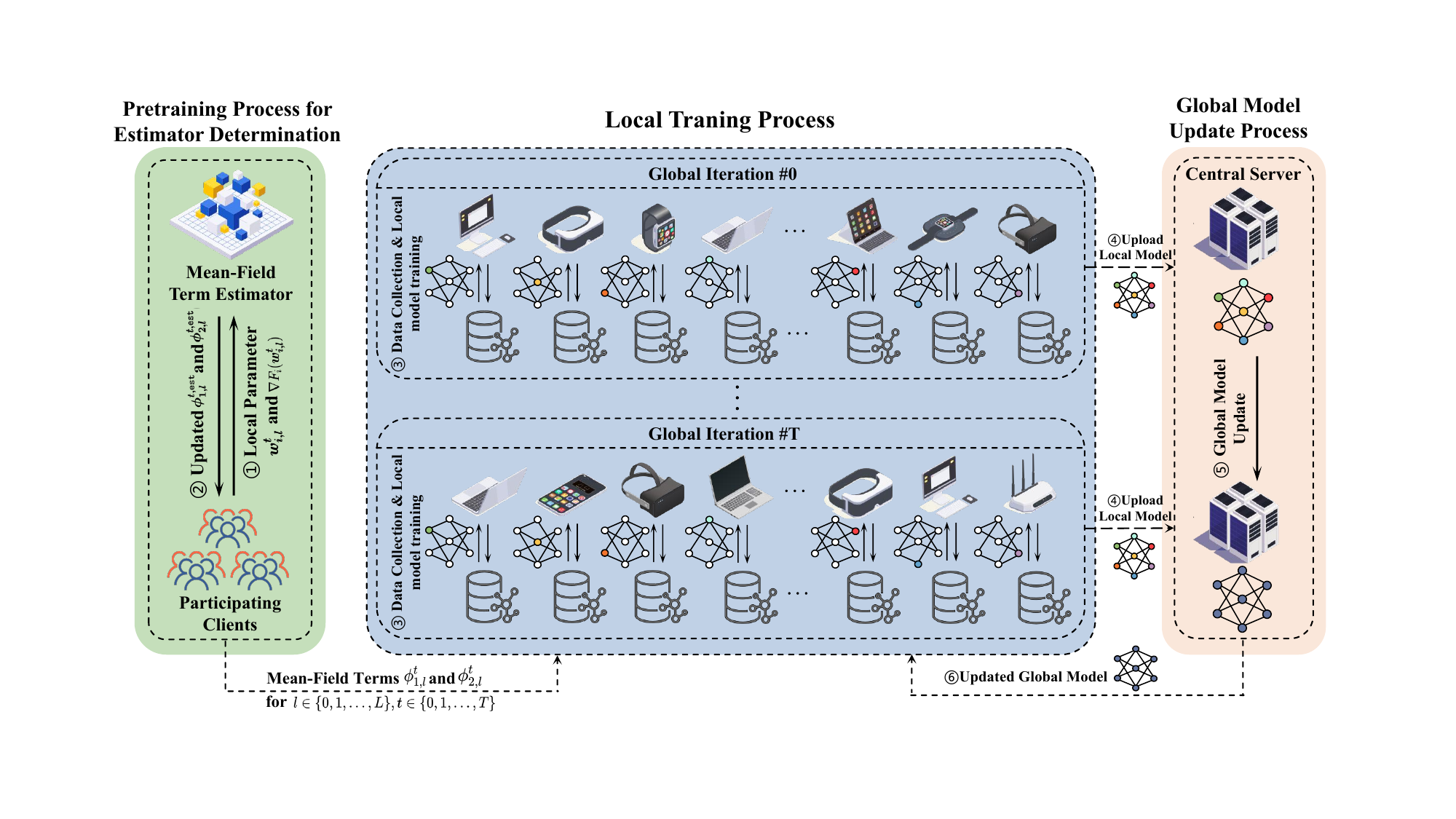}}
\caption{The federated learning framework with aggregated gradients.}
\label{framework}
\vspace{-12pt}
\end{figure*}

The rest of this paper is organized as follows. We review the related work in Section \ref{related work}. System model and problem formulation are given in Section \ref{system_models_and_problem_formulation}. In Section \ref{analysis_of_adaptive_learning_rate}, an analysis of the adaptive learning rate is presented. We provide the convergence analysis in Section \ref{Convergence_analysis}. Experimental results are shown in Section \ref{Numerical_Experiments}. We conclude this paper in Section \ref{Conclusion}.

\section{Related Work} \label{related work}
In this section, we briefly review the related literature on adaptive optimization in federated learning from the aspect of hyperparameter update, client selection mechanism design, and asynchronous aggregation.

\subsection{Adaptive Hyperparameter Update in FL} Zhang \emph{et al.} \cite{zhang2021adaptive} propose an adaptive FL algorithm, named Adaptive-B, that obtains the most suitable batch size effectively through the Deep Reinforcement Learning (DRL) method. Ma \emph{et al.} \cite{ma2021adaptive} study the relationship between the batch size and learning rate and formulate a scheme to adaptively adjust the batch size with a scaled learning rate to reduce devices' waiting time and improve the convergence rate. Wu \emph{et al.} \cite{wu2021fast} propose FedAdp to assign different weights for updating the global model based on node contribution adaptively. Wu \emph{et al.} \cite{wu2022adaptive} propose Fadam that adaptively adjusts the local model gradient based on the first and second-order momentum of the historical gradient to reduce overfitting and fluctuation. Zhang \emph{et al.} \cite{zhang2023timelyfl} propose TimelyFL that adaptively adjusts the workload (i.e., the local epoch number) for each client to finish the local training and model upload. Wu \emph{et al.} \cite{wu2023hiflash} propose HiFlash that achieves the trade-off between efficient communication and model accuracy by designing an adaptive edge staleness threshold for clients.

\subsection{Adaptive Client Selection Mechanism in FL} 
Li \emph{et al.} \cite{li2021fedsae} introduce FedSAE that automatically adjusts the training task of clients and converts the training loss of each client into a selected probability, based on which the central server chooses the optimal participants. Luo \emph{et al.} \cite{luo2022tackling} design an adaptive algorithm that optimizes the client sampling probability to tackle both system and statistical heterogeneity and minimize wall-clock convergence time. Huang \emph{et al.} \cite{huang2022tackling} introduce a strategy named FEDEMD that adaptively selects clients based on Wasserstein distance between the local and global data distributions. Huang \emph{et al.} \cite{huang2023active} propose an ACFL scheme that selects clients based on several active learning (AL) metrics to mitigate the data heterogeneity on model performance. Xu \emph{et al.} \cite{xu2023enhancing} introduce the Ada-FedSemi system that adaptively determines the participating ratio and confidence threshold of pseudo-labels with a multi-armed bandit (MAB) based online algorithm. Jiang \emph{et al.} \cite{jiang2023heterogeneity} propose a heterogeneity-aware FL framework FedCG that effectively alleviates the communication burden and mitigates the straggler effect by adaptive client selection and gradient compression.

\subsection{Adaptive Asynchronous Aggregation in FL} 
Liu \emph{et al.} \cite{liu2021adaptive} design AAFL mechanism, which adaptively determines the optimal fraction of clients participating in global aggregation based on resource constraints. Liu \emph{et al.} \cite{liu2022fed2a} introduce Fed2A which allows clients to upload shallow and deep layers of DNNs adaptively to improve their performance in a heterogeneous and asynchronous environment. Wang \emph{et al.} \cite{wang2022asyncfeded} propose AsyncFedED, considering the staleness of the arrived gradients measured by the Euclidean distance between the stale and the current global model, as well as the number of local epochs. Zhang \emph{et al.} \cite{zhang2023fedmds} propose a semi-asynchronous clustered FL framework, named FedMDS, which adopts a clustered strategy and synchronous trigger mechanism to alleviate the model staleness problem. Dun \emph{et al.} \cite{dun2023efficient} propose a novel asynchronous FL framework AsyncDrop that leverages the dropout regularization approach to address the device heterogeneity. Hu \emph{et al.} \cite{hu2023scheduling} introduce a scheduling policy, jointly considering the channel quality and data distribution, that achieves periodic aggregation and fast convergence in the asynchronous FL framework with~stragglers.

In the existing literature, a significant shortfall in the domain of adaptive FL pertains to the absence of rigorously theoretical analyses for the closed-form expression of the adaptive parameters and quantifying the model divergence. Our contribution contrasts with the prevailing research by effectively bridging this gap. We tackle the challenges of data heterogeneity and client drift while maintaining model accuracy by dynamically adjusting the learning rates. By incorporating a penalty term that quantifies the divergence between local and global model parameters, we derive a closed-form solution for the adaptive learning rate, thereby enhancing overall model efficacy. To underscore the robustness and broad applicability of FedAgg, we conduct comprehensive theoretical convergence analyses.

\begin{table}[t]
\setlength{\abovecaptionskip}{0cm} 
\caption{Key Notations and Corresponding Meanings}
\begin{center}
\renewcommand\arraystretch{1.2}
\begin{tabular}{p{0.5in}p{2.65in}}
\toprule[0.8pt]
\makecell[l]{\textbf{Symbol}}&\makecell[c]{\textbf{Description}}\\
\midrule[0.8pt]
$N$ & number of participating clients   \\

$L$ & total number of local training epoch  \\

$T$ & total number of global training iteration  \\

$\mathcal{D}_{i}$, $D_{i}$ & private data on client $i$ and corresponding datasize   \\

$\boldsymbol{w_{i,l}^{t}}$ & local model parameter of client $i$ at the $l$-th local epoch of global iteration $t$    \\

$\boldsymbol{\bar w^{t}}$ & global model parameter at $t$-th global iteration   \\

$F_{i}(\boldsymbol{w_{i,l}^{t}})$ & the local loss function of client $i$  \\

$F(\boldsymbol{w})$ & the global loss function  \\

$\nabla F$ & the gradient of loss function  \\

$\eta_{i,l}^{t}$ & learning rate of client $i$ at the $l$-th local epoch of global iteration $t$  \\ 

$\boldsymbol{\phi_{1,l}^{t}}$, $\boldsymbol{\phi_{2,l}^{t}}$ & mean-field terms for estimating average gradient and parameter of all clients \\

\bottomrule[0.8pt]
\end{tabular}
\label{tab1}
\end{center}
\vspace{-8pt}
\end{table}

\section{System Model and Problem Formulation} \label{system_models_and_problem_formulation}
In this section, we first introduce the preliminaries regarding the standard FL model in Section \ref{standard_fl_model}. Then, we formulate our FL optimization problem by considering the local model deviation of each client in Section \ref{problem_formulation}. The workflow of our proposed FL framework with aggregated gradient is presented in Fig.~\ref{framework} and the key notations and corresponding meanings we use throughout the paper are summarized in Table~\ref{tab1}. 

\subsection{Standard Federated Learning Model} \label{standard_fl_model}
Federated learning represents a decentralized machine learning paradigm explicitly designed to address the challenges of privacy concerns. Within the FL framework, a global model is collaboratively trained by substantial clients with locally collected data. Given $N$ geographically distributed and heterogeneous clients (i.e. $S \!=\! \{1,2, \ldots, N\}$) willing to participate in the FL training, and each client $i \in \{1, 2, \ldots, N\}$ utilizes their private local data $\mathcal{D}_{i}$ with datasize $D_{i}$ to perform model training. At global training iteration $t \!\in\! \{0,1, \ldots, T\}$, each client $i$ performs $L$ ($L \!\geq\! 1$) epochs of mini-batch SGD training to update its local model parameter parallelly as follows:
\begin{align} \label{local_parameter_update}
\boldsymbol{w_{i,l+1}^{t}}=\boldsymbol{w_{i,l}^{t}}-\eta_{i} \nabla F_{i}(\boldsymbol{w_{i,l}^{t}}), 
\end{align} 
where $\eta_{i}$ represents the learning rate of client $i$. In classic federated algorithms, such as FedAvg, $\eta_{i}$ is typically defaulted as a constant. We respectively denote $\boldsymbol{w_{i,l}^{t}}$ and $\nabla F_{i}(\boldsymbol{w_{i,l}^{t}})$ as the local parameter and gradient of client $i$ at the $l$-th local epoch of global iteration $t$, where $l \!\in\! \{0,1,  \ldots, L\}$. The central server averages the local models to generate the updated global parameter as soon as receiving all local parameters by:
\begin{align} \label{local_model_aggregation}
\boldsymbol{\bar w^{t}}=\sum\nolimits_{i=1}^{N} \frac{D_{i}}{\sum\nolimits_{j=1}^{N} D_{j}} \boldsymbol{w_{i,L}^{t-1}}.
\end{align}

At the onset of $t$-th iteration, we default $\boldsymbol{w_{i,0}^{t}} \!=\! \boldsymbol{\bar w^{t}}$. After local model aggregation, the central server dispatches the updated global model back to all clients for the next iteration's training. The goal of the FL method is to find optimal global parameter $\boldsymbol{w^{*}}$ to minimize the global loss function $F(\boldsymbol{w})$:
\begin{align} \label{fl_goal}
\min_{\boldsymbol{w}}  F(\boldsymbol{w})=\sum\nolimits_{i=1}^{N} \frac{D_{i}}{\sum\nolimits_{j=1}^{N} D_{j}} F_{i}(\boldsymbol{w}).
\end{align}

For further demonstration and theoretical analysis, we introduce the following assumptions and lemmas which are widely used in federated optimization problems \cite{mcmahan2017communication, mitra2021achieving, li2023anarchic, bottou2018optimization}.
\begin{assumption} \label{assumption_smoothness}
($\beta$-Lipschitz Smoothness) $F_{i}(\boldsymbol{w})$ satisfies $\beta$-Lipschitz smoothness for each clients $i \!\in\! \{0,1, \ldots, N\}$ with any two parameter vectors $\boldsymbol{w}$ and $\boldsymbol{w}^{\prime}$, we have: 
\begin{align}
&\|\nabla{F_{i}(\boldsymbol{w}}) \!-\! \nabla{F_{i}(\boldsymbol{w}^{\prime}})\| \!\leq\! \beta\|\boldsymbol{w} \!-\! \boldsymbol{w}^{\prime}\|,  \\
F(\boldsymbol{w}) &\!-\! F(\boldsymbol{w}^{\prime}) \!\leq\! \nabla{F(\boldsymbol{w}^{\prime})}^\top(\boldsymbol{w}\!-\!\boldsymbol{w}^{\prime}) \!+\!\frac{\beta}{2} \|\boldsymbol{w}\!-\!\boldsymbol{w}^{\prime}\|_{2}^{2}.
\end{align}
\end{assumption}

\begin{assumption}\label{bounded_Gradients}
(Bounded Local Gradient) With the continuously differentiable local loss function $F_{i}(\boldsymbol{w_{i,l}^{t}})$, the local gradient $\nabla F_{i}(\boldsymbol{w})$ attains $P$-bound gradients for any client $i \!\in\! \\ \{1, 2, \ldots,  N\}$, we have: $\|\nabla{F_{i} (\boldsymbol{w_{i,l}^{t}})}\|_{2} \!\le\! P$.
\end{assumption}

\begin{assumption} \label{assumption_convexity}
($\psi$-Strong Convexity) Global loss function $F(\boldsymbol{w})$ is $\psi$-strong convex with any $\boldsymbol{w}, \boldsymbol{w}^{\prime} \in \mathbb{R}^{d}$: 
\begin{align}
\frac{\psi}{2}\|\boldsymbol{w} - \boldsymbol{w^{*}}\|_{2}^{2} \leq F(\boldsymbol{w}) - F(\boldsymbol{w}^{*}). 
\end{align}
\end{assumption}

\begin{assumption} \label{assumption_subgradient}
[Subgradient for Non-Convexity] Global loss function $F(\boldsymbol{w})$ admits a unique subgradient $ g \!\in\! \mathbb{R}^{d}$ for any model parameter vector $\boldsymbol{w}, \boldsymbol{w}^{\prime} \!\in\! \mathbb{R}^{d}$, it satisfies, 
\begin{align}
F(\boldsymbol{w}) - F(\boldsymbol{w}^{\prime}) \geq g^\top (\boldsymbol{w} - \boldsymbol{w}^{\prime}),
\end{align}
\end{assumption}

\begin{assumption}\label{assumption_bounded_parameter}
(Bounded Local Model Parameter) For client $ i \!\in\! \{1, 2, \ldots, N\}$, the $\ell_{2}$-norm of the local model parameter $\|\boldsymbol{w_{i,l}^{t}}\|_{2}$ is upper bounded by constant $Q$, i.e., $ \|\boldsymbol{w_{i,l}^{t}}\| \!\le\! Q$.
\end{assumption}

\begin{lemma} \label{PL_inequality}
[Polyak-\L{}ojasiewicz Inequality] With the optimal global parameter $\boldsymbol{w}^{*}$ and suppose that the global loss function $F(\boldsymbol{w})$ satisfies $\beta$-Lipschitz continuous and Polyak-\L{}ojasiewicz condition. Then, for any $\mu \!>\! 0$ and $\boldsymbol{w} \!\in\! \mathbb{R}^{d}$, we hold:  
\begin{align}
\frac{1}{2}\|\nabla{F(\boldsymbol{w})}\|_{2}^{2} \!\geq\! \mu(F(\boldsymbol{w}) \!-\! F(\boldsymbol{w}^{*})). 
\end{align}
\end{lemma}

\subsection{Problem Formulation} \label{problem_formulation}
The adaptive learning rate mechanism is crucial and efficient in deep learning for addressing large-scale optimization problems \cite{zhou2018adashift}, as illustrated in Fig.~(\ref{beta}). It is important to recognize that conventional FL methods generally cause client drifting during local updates, resulting in slow and unstable convergence \cite{karimireddy2020scaffold}. To alleviate the adverse effects of client drifting brought by Non-IID data and heterogeneous clients while achieving fast model convergence in FL systems, we adopt an adaptive learning rate $\eta_{i,l}^{t}$ and introduce an aggregated gradient term $\frac{1}{N} \sum\nolimits_{i=1}^{N} \nabla{F_{i}(\boldsymbol{w_{i,l}^{t}})}$ for each client in local model update:
\begin{align}\label{local_model_update_with_aggregated_gradient}
\boldsymbol{w_{i,l+1}^{t}}=\boldsymbol{w_{i,l}^{t}} - \eta_{i,l}^{t} \left(\frac{1}{N} \sum\nolimits_{i=1}^{N} \nabla{F_{i}(\boldsymbol{w_{i,l}^{t}})} \right),
\end{align} 

To accelerate the FL model convergence rate and mitigate pronounced volatility throughout the FL training, each client aims to cautiously design an adaptive learning rate $\eta_{i,l}^{t}$ at each iteration. We adopt an $\ell_{2}$-norm based penalty term to measure the deviation between the local model parameter of client $i$ and the average local model parameter at $l$-th local epoch. The metric capitalizes on its susceptibility to outliers and aptitude in quantifying variable discrepancies across multi-dimensional spaces, providing a distinct advantage over the conventionally utilized $\ell_{1}$-norm and cosine similarity metrics \cite{pandit2011comparative}. Hence, subject to the constrain in Eq.~(\ref{local_model_update_with_aggregated_gradient}), the objective of each client $i$ is to design an adaptive learning rate $\eta_{i,l}^{t}$ at each iteration to minimize its training cost $U_{i}(T)$ including the penalty term for quantifying the local model deviation and the learning rate $\eta_{i,l}^{t}$ to avoid significant fluctuation and gradient explosion during local model training \cite{luo2019adaptive}:
\begin{align} 
\!\!\!\! U_{i}(T) \!= &  \min_{\substack{\eta_{i,l}^{t}, l \in \{0,1, \ldots, L\},\\ t \in \{0,1, \ldots, T\}}}  \sum\nolimits_{t=0}^{T}\! \sum\nolimits_{l=0}^{L} \alpha (\eta_{i,l}^{t})^{2}\!+\!(1\!-\!\alpha) \nonumber \\ 
&  \times\! (\boldsymbol{w_{i,l}^{t}}\!-\!\frac{1}{N}\! \sum\nolimits_{j=1}^{N} \!\!\boldsymbol{w_{j,l}^{t}})^\top \! (\boldsymbol{w_{i,l}^{t}}\!-\!\frac{1}{N}\! \sum\nolimits_{j=1}^{N} \!\!\boldsymbol{w_{j,l}^{t}}), \!\!\!\!   \label{objective}  \\
s.t. &  \quad   \boldsymbol{w_{i,l+1}^{t}}=\boldsymbol{w_{i,l}^{t}} - \eta_{i,l}^{t} \left(\frac{1}{N} \sum\nolimits_{i=1}^{N} \nabla{F_{i}(\boldsymbol{w_{i,l}^{t}})} \right), \tag{9}
\end{align} 
where the aggregation weight $\alpha \!\in\! [0, \!1]$ is predefined to quantify the extent of focus between model parameter divergence and the adaptive learning rate adjustment in objective function $U_{i}(T)$. An elevated value of $\alpha$ signifies the pronounced impact of local parameter discrepancies on the FL system.


\section{Analysis of Adaptive Learning Rate} \label{analysis_of_adaptive_learning_rate}
In this section, we first derive the optimal decentralized adaptive learning rate $\eta_{i,l}^{t}$ in closed-form for each client by introducing mean-field terms $\boldsymbol{\phi_{1,l}^{t}}$ and $\boldsymbol{\phi_{2,l}^{t}}$ to estimate the average local parameters and gradients respectively. Then, we demonstrate the existence and convergence of mean-field terms. Finally, an iterative algorithm is proposed for their determination.

\subsection{Decentralized Learning Rate For Each Client}\label{Decentralized Learning Rate For Each Client}

From the objective function $U_{i}(T)$ in Eqs.~(\ref{local_model_update_with_aggregated_gradient})-(\ref{objective}), we observe that the learning rate $\eta_{i,l}^{t}$ of each client is affected by other clients' local parameters $\boldsymbol{w_{j,l}^{t}}$ and $\nabla{F_{j}(\boldsymbol{w_{j,l}^{t})}}$, $j \!\in\! S \!\setminus\! \{i\}$.~At each local epoch, as there is no information exchange among clients and central server, client $i$ can not access other participants' local information $\boldsymbol{w_{j,l}^{t}}$ and $\nabla{F_{j}(\boldsymbol{w_{j,l}^{t})}}$, which makes this multi-client joint adaptive learning rate design over time challenging.~To tackle the above challenge, we introduce two mean-field terms to estimate the average local parameters and gradients, based on which, the decentralized adaptive learning rate of each client can be derived without requiring other clients' local information. The definition of mean-field terms is given as follows.
\begin{definition}\label{definition_mean_field_terms}
(Mean-Field Terms) To figure out the optimal decentralized learning rate for each client, we introduce two mean-field terms to respectively estimate the average local gradients and parameters of all clients at the $l$-th local epoch of global iteration $t$, where $t \in\{0,1, \ldots, T\}$, $l \in\{0,1, \ldots, L\}$:
\begin{align} 
\boldsymbol{\phi_{1,l}^{t}} &= \frac{1}{N}\sum\nolimits_{i=1}^{N} \nabla{F_{i} (\boldsymbol{w_{i,l}^{t}})}, \label{phi1} \\
\boldsymbol{\phi_{2,l}^{t}} &= \frac{1}{N}\sum\nolimits_{i=1}^{N} \boldsymbol{w_{i,l}^{t}}. \label{phi2}
\end{align}
\end{definition}

From the mathematical point of view, terms $\boldsymbol{\phi_{1,l}^{t}}$ and $\boldsymbol{\phi_{2,l}^{t}}$ are given functions in the our FL optimization problem. Based on Definition \ref{definition_mean_field_terms}, the deviation between the local parameter of client $i$ and the average local parameter can be further refined as $\boldsymbol{w_{i,l}^{t}} - \boldsymbol{\phi_{2,l}^{t}}$. Accordingly, the optimization problem in Eqs.~(\ref{local_model_update_with_aggregated_gradient})-(\ref{objective}) can be reformulated as follows:
\begin{align} 
\Tilde{U}_{i}(T) =& \min_{\substack{\eta_{i,l}^{t}}} \sum\nolimits_{t=0}^{T} \sum\nolimits_{l=0}^{L}   \alpha(\eta_{i,l}^{t})^{2} \nonumber \\
&  + (1 - \alpha) (\boldsymbol{w_{i,l}^{t}} - \boldsymbol{\phi_{2,l}^{t}})^\top (\boldsymbol{w_{i,l}^{t}} - \boldsymbol{\phi_{2,l}^{t}}),   \label{objective_mean_field} \\ 
s.t. &  \quad \boldsymbol{w_{i,l+1}^{t}}=\boldsymbol{w_{i,l}^{t}}-\eta_{i,l}^{t}  \boldsymbol{\phi_{1,l}^{t}}. \label{local_model_update_mean_field}
\end{align} 

The optimization problem in Eqs. (\ref{objective_mean_field})-(\ref{local_model_update_mean_field}) can be characterized as a discrete-time linear quadratic optimal control problem. The learning rate at $t$-th iteration exerts a pivotal influence on the subsequent iterations. Leveraging this insight, by constructing the Hamilton function, we can derive the closed-form expression of adaptive learning rate $\eta_{i,l}^{t}$ as summarized in the following theorem.
\begin{theorem} \label{theorem_optimal_lr}
(Optimal Adaptive Learning Rate) Given the mean-field terms $ \boldsymbol{\phi_{1,l}^{t}}$ and $ \boldsymbol{\phi_{2,l}^{t}}$, the optimal adaptive learning rate of client $i \!\in\! \{1, 2, \ldots, N\}$ at $l$-th local epoch of global iteration $t$, $l \!\in\! \{0,1, \ldots, L\}$, $t \!\in\! \{0,1, \ldots, T\}$ is derived~as:
\begin{align}\label{optimal_learning_rate}
\eta_{i,l}^{t} =& \frac{1 - \alpha}{\alpha}(\boldsymbol{\phi_{1,l}^{t}})^\top \times\! \left[(L - l)\boldsymbol{w_{i,l}^{t}} \right. \nonumber \\
& \left. - \sum\nolimits_{r=l}^{L} (L - r)\eta_{i,r}^{t}\boldsymbol{\phi_{1,r}^{t}} -\sum\nolimits_{k=l+1}^{L} \boldsymbol{\phi_{2,k}^{t}}\right],
\end{align}
with $ \eta_{i,L}^{t}\!=\!0$.
\end{theorem}

\begin{proof}
Based on the objective function presented in Eq.~(\ref{objective_mean_field}) and mean-field terms $\boldsymbol{\phi_{1,l}^{t}}$ and $ \boldsymbol{\phi_{2,l}^{t}}$, we construct the discrete-time Hamilton function as follows:
\begin{align} \label{Hamilton}
H(l) =& \ (1 - \alpha )(\boldsymbol{w_{i,l}^{t}} - \boldsymbol{\phi_{2,l}^{t}})^\top (\boldsymbol{w_{i,l}^{t}} - \boldsymbol{\phi_{2,l}^{t}}) \nonumber \\
&+ \alpha (\eta_{i,l}^{t})^{2} - \boldsymbol{\lambda(l + 1)}^\top \eta_{i,l}^{t}\boldsymbol{\phi_{1,l}^{t}},
\end{align} 
where $\boldsymbol{\phi_{1,l}^{t}}$ and $\boldsymbol{\phi_{2,l}^{t}}$ are given functions from Definition~\ref{definition_mean_field_terms}, and not affected by the learning rate $\eta_{i,l}^{t}$. According to the properties of the discrete-time Hamilton function in \cite{di2012discrete}, to obtain the expression of $\eta_{i,l}^{t}$, we firstly consider the first-order derivation constraint on $ \eta_{i,l}^{t}$, which is viewed as the control vector in Eq.~(\ref{Hamilton}). For all time instants, the minimum hold by calculating the first derivative of Eq.~(\ref{Hamilton}) concerning $\eta_{i,l}^{t}$, i.e., $\partial H(t)/\partial(\eta_{i,l}^{t}) \!=\! 2\alpha \eta_{i,l}^{t} - \boldsymbol{\lambda(l + 1)}^\top \boldsymbol{\phi_{1,l}^{t}} = 0$. Thus, we obtain the expression of the learning rate $\eta_{i,l}^{t}$ as follows:
\begin{align} \label{eta}
\eta_{i,l}^{t} = \frac{1}{2 \alpha}(\boldsymbol{\phi_{1,l}^{t}})^\top \boldsymbol{\lambda(l + 1)}.
\end{align}

Through calculating the second-order derivative of the Hamilton function in Eq.~(\ref{Hamilton}),  i.e., $\partial^{2} H(t) / \partial (\eta_{i,l}^{t})^{2} \!=\! 2 \alpha \!>\! 0$, to validate the existence of the minimum, we obtain the adaptive learning rate $\eta_{i,l}^{t}$ that minimizes the objective function. According to the differential property in \cite{qi2021linear} concerning the impulse vector $\boldsymbol{\lambda}$ and the state vector $\boldsymbol{w}$ in Eq.~(\ref{Hamilton}), we have the following iterative equation:
\begin{align} \label{lambda_difference}
\boldsymbol{\lambda(l+1)}\!-\!\boldsymbol{\lambda(l)} \!=\! -\frac{\partial H(t)}{\partial(\boldsymbol{w_{i,l}^{t}})} \!=\! -2(1\!-\!\alpha)(\boldsymbol{w_{i,l}^{t}}\!-\!\boldsymbol{\phi_{2,l}^{t}}).
\end{align}

Based on the boundary condition in \cite{di2012discrete} for the final value at $L$-th local epoch of the impulse vector $\boldsymbol{\lambda}$, we have: $\boldsymbol{\lambda(L)}=\frac{\partial S(\boldsymbol{w_{i,L}^{t}})}{\partial (\boldsymbol{w_{i,L}^{t}})}= 2(1-\alpha)(\boldsymbol{w_{i,L}^{t}}-\boldsymbol{\phi_{2,L}^{t}})$, where $S(\cdot)$ is a weighting function of the state at the $L$-th local epoch and follows the equation $S(\boldsymbol{w_{i, L}^{t}}) \!=\! (1 \!-\! \alpha)(\boldsymbol{w_{i, L}^{t}} \!-\! \boldsymbol{\phi_{2, L}^{t}})^\top(\boldsymbol{w_{i, L}^{t}} \!-\! \boldsymbol{\phi_{2, L}^{t}})$. By iteratively aggregating the formulations in Eq.~(\ref{lambda_difference}) with $l \!\in\! \{0,1, \ldots, L-1\}$, the expression of $\boldsymbol{\lambda(l+1)}$ is given by:
\begin{align} \label{lambda}
\!\!\!\! \boldsymbol{\lambda(l \!+\! 1)} =&\ 2 (1 \!-\! \alpha) \times \left[(L -l) \boldsymbol{w_{i,l+1}^{t}} \right. \nonumber \\
& \left. -\! \sum\nolimits_{r=l+1}^{L}\!(L - r)\eta_{i,r}^{t} \boldsymbol{\phi_{1,r}^{t}} \!-\! \sum\nolimits_{k=l+1}^{L} \! \boldsymbol{\phi_{2,k}^{t}}\right]. \!\!
\end{align}

Then, combining the expression of $\eta_{i,l}^{t}$ in Eq.~(\ref{eta}) and the impulse vector $\boldsymbol{\lambda(l+1)}$ in Eq.~(\ref{lambda}), we finalize the optimal learning rate $\eta_{i,l}^{t}$ as shown in Eq.~(\ref{optimal_learning_rate}). \end{proof}

From Theorem~\ref{theorem_optimal_lr}, we observe that the mathematical formula of $\eta_{i,l}^{t}$ at the $l$-th local epoch of global iteration $t$ is associated with the learning rate $\eta_{i,k}^{t}$, $k \!\in\! \{l, l + 1, \ldots, L\}$. By backward induction, we prove that $\eta_{i,l}^{t}$ is solvable and summarized in the following proposition.  
\begin{proposition} \label{proposition_lr_linear_combination}
$ \eta_{i,l}^{t}$ is solvable and can be expressed as a linear combination of $\boldsymbol{w_{i,l}^{t}}$, $ \boldsymbol{\phi_{1,l}^{t}}$ and $ \boldsymbol{\phi_{2,l}^{t}}$, where $ l \in \{0, 1, \\ \ldots, L\}$, $ t \!\in\! \{0,1, \ldots, T\}$.
\end{proposition}

The detailed proof of Proposition~\ref{proposition_lr_linear_combination} is provided in Appendix~A of the supplementary file. Theorem~\ref{theorem_optimal_lr} illustrates that the adaptive learning rate of client $i$ in Eq.~(\ref{optimal_learning_rate}) is only associated with its local information $\boldsymbol{w_{i,l}^{t}}$ and mean-field terms $\boldsymbol{\phi_{1,l}^{t}}$ and $ \boldsymbol{\phi_{2,l}^{t}}$. In the following, we will show how to obtain the mean-field terms.

\subsection{Update of Mean-Field Estimators for the Adaptive Learning Rate Finalization}
Note that mean-field estimators $\boldsymbol{\phi_{1,l}^{t}}$ and $\boldsymbol{\phi_{2,l}^{t}}$ are affected by the local gradient $ \nabla{F_{i}(\boldsymbol{w_{i,l}^{t})}}$ and local parameter $\boldsymbol{w_{i,l}^{t}}$ of all clients, which will inversely affect the determination of the value of $\nabla{F_{i}(\boldsymbol{w_{i,l}^{t})}}$ and $ \boldsymbol{w_{i,l}^{t}}$. Given the optimal adaptive learning rate $ \eta_{i,l}^{t}$ in Theorem \ref{theorem_optimal_lr}, we can obtain the solutions for the mean-field terms $\boldsymbol{\phi_{1,l}^{t}}$ and $ \boldsymbol{\phi_{2,l}^{t}}$ by finding the fixed~point.
\begin{theorem} \label{theorem_fixed_point}
There exists a fixed point for the mean-field terms $\boldsymbol{\phi_{1,l}^{t}}$ and $\boldsymbol{\phi_{2,l}^{t}}$ in Definition~\ref{definition_mean_field_terms}, where $t \!\in\! \{0, 1, \ldots, T\}, \\ l \!\in\! \{0, 1, \ldots, L\}\})$.
\end{theorem}

\begin{algorithm}[t]
\caption{Pretraining for estimating $\boldsymbol{\phi_{1,l}^{t}}$ and $\boldsymbol{\phi_{2,l}^{t}}$} 
\begin{algorithmic}[1]
\State {\bf Input:} 
$L$, $T$, $\epsilon_{1} \!=\! \epsilon_{2} \!=\! 1$, $\epsilon \!=\! 10^{-3}$, $j \!=\! 1$, initial value for $\boldsymbol{w_{i,0}^{t}}$, $\boldsymbol{\phi_{1,l}^{t}}\!=\!\boldsymbol{\phi_{1,l}^{t,\texttt{est}}}(0) \!\ge\! 0$, $\boldsymbol{\phi_{2,l}^{t}}\!=\!\boldsymbol{\phi_{2,l}^{t,\texttt{est}}}(0) \!\ge\! 0$ and learning rate $\eta_{i,l}^{t} \!\in\! (0, 1)$.
\For{$t=0$ to $T$}
    \While{$\epsilon_{1} \!>\! \epsilon$ and $\epsilon_{2} \!>\! \epsilon$}
        \For{each client $i \!\in\! \{0,1, \ldots, N\}$}
            \For{$l=1$ to $L-1$}
            \State Compute $\eta_{i,l}^{t}$, $\boldsymbol{w_{i,l}^{t}}$ and $\nabla \! F_{i} (\boldsymbol{w_{i,l}^{t}})$
            \EndFor
        \EndFor
            \For{$l=0$ to $L-1$}
            \State $\boldsymbol{\phi_{1,l}^{t,\texttt{est}}}(j) \!=\! \frac{1}{N} \!\! \sum_{i=1}^{N}\!\!\! \nabla \! F_{i} (\boldsymbol{w_{i,l}^{t})}$, $\!\boldsymbol{\phi_{1,l}^{t}} \!=\! \boldsymbol{\phi_{1,l}^{t,\texttt{est}}}(j)$
            \State $\boldsymbol{\phi_{2,l}^{t,\texttt{est}}}(j)=\frac{1}{N}\sum_{i=1}^{N} \boldsymbol{w_{i,l}^{t}}$, $\boldsymbol{\phi_{2,l}^{t}}=\boldsymbol{\phi_{2,l}^{t,\texttt{est}}}(j)$
            \State $\epsilon_{1}'=\boldsymbol{\phi_{1,l}^{t,\texttt{est}}}(j)-\boldsymbol{\phi_{1,l}^{t,\texttt{est}}}(j-1)$
            \State $\epsilon_{2}'=\boldsymbol{\phi_{2,l}^{t,\texttt{est}}}(j)-\boldsymbol{\phi_{2,l}^{t,\texttt{est}}}(j-1)$
            \EndFor
            \State $\epsilon_{1}=\sum_{l=0}^{L} \epsilon_{1}'$, $\epsilon_{2}=\sum_{l=0}^{L} \epsilon_{2}'$, $j=j+1$
    \EndWhile
\EndFor
\State \Return Mean-field terms $\boldsymbol{\phi_{1,l}^{t}}$ and $\boldsymbol{\phi_{2,l}^{t}}$, $t \!\in\! \{0,1, \ldots, T\}$.
\end{algorithmic}
\label{alg1}
\end{algorithm}

\begin{proof} 
By iteratively aggregating the equation in Eq.~(\ref{local_model_update_with_aggregated_gradient}) in time instants $t \!\in\! \{0,1, \ldots, T\}$, the refined expression of $\boldsymbol{w_{i,l}^{t}}$ in Eq.~(\ref{local_model_update_with_aggregated_gradient}) is given by
\begin{align}\label{fresh_local_update}
\!\!\!\!\! \boldsymbol{w_{i,l}^{t}} \!=&\ \boldsymbol{w_{i,0}^{t}}-\sum\nolimits_{p=0}^{l} \eta_{i,p}^{t} (\frac{1}{N}\sum\nolimits_{j=1}^{N} \nabla{F_{j} (\boldsymbol{w_{j,p}^{t}})}) \nonumber \\ 
=&\  \boldsymbol{w_{i,0}^{t}} \!-\! \sum\nolimits_{p=0}^{l} (\sum\nolimits_{i=1}^{N}\!\! \nabla{F_{i} (\boldsymbol{w_{i,p}^{t}})})^\top (\sum\nolimits_{j=1}^{N} \!\! \nabla{F_{j} (\boldsymbol{w_{j,p}^{t}})}) \nonumber \\ 
& \times \left[N (L \!-\! p) \boldsymbol{w_{i,p}^{t}} - \sum\nolimits_{k=p+1}^{L}\! \sum\nolimits_{i=1}^{N}\boldsymbol{w_{i,k}^{t}} \right. \nonumber \\
& - \left. \sum\nolimits_{r=p}^{L} \sum\nolimits_{q=1}^{N} \! (L \!-\! r) \eta_{i,r}^{t}\nabla F_{q} (\boldsymbol{w_{q,r}^{t}}) \right] \!\times\! \frac{1 \!-\! \alpha}{\alpha N^{3}}. 
\end{align}

Then, for any client $i \!\in\! \{1, 2, \ldots, N\}$, we substitute $\boldsymbol{\phi_{1,l}^{t}} \!=\! \frac{1}{N} \! \sum_{i=1}^{N} \! \nabla F_{i} (\boldsymbol{w_{i,l}^{t}}) $ and $ \boldsymbol{\phi_{2,l}^{t}} \!=\! \frac{1}{N} \! \sum_{i=1}^{N} \! \boldsymbol{w_{i,l}^{t}}$, where $ t \!\in\! \{0,1, \\ \ldots,  T\}$, $l \!\in\! \{0,1, \ldots, L\}$ into Eq.~(\ref{fresh_local_update}), we observe that $\boldsymbol{w_{i,l}^{t}}$ and $ \nabla \!F_{i} (\boldsymbol{w_{i,l}^{t}})$ of client $i$ at the $l$-th local epoch of global iteration $t$ are a function of $ \{\boldsymbol{w_{i,l}^{t}}, \nabla \! F_{i} (\boldsymbol{w_{i,l}^{t}}) \mid t \in \{0,1, \ldots, \\ T\}, l \!\in\! \{0,1, \ldots, L\}\}$ of all clients. Define the following function as a mapping from $ \{\boldsymbol{w_{i,l}^{t}}, \nabla \! F_{i} (\boldsymbol{w_{i,l}^{t}}) \mid t \in \{0,1, \ldots, \\ T\}, l \!\in\! \{0,1, \ldots, L\}\}$ to client $i$'s local parameter $ \boldsymbol{w_{i,l}^{t}}$ and local gradient $ \nabla \! F_{i} (\boldsymbol{w_{i,l}^{t}})$ in Eq.~(\ref{fresh_local_update}) at global iteration $t$, i.e., $ (\boldsymbol{w_{i,l}^{t}}, \nabla{F_{i} (\boldsymbol{w_{i,l}^{t}})}) = \Gamma_{l}^{t}(\{\boldsymbol{w_{i,l}^{t}}, \nabla F_{i} (\boldsymbol{w_{i,l}^{t}}) \!\mid t \in \{0, 1, \ldots,T\}, \\ l \!\in\! \{0,1, \ldots, L\}\})$. Thus, to summarize any possible mapping $ \Gamma_{l}^{t}$, we define the following vector function as a mapping from $\{\boldsymbol{w_{i,l}^{t}}, \nabla \! F_{i} (\boldsymbol{w_{i,l}^{t}}) \!\mid\! t \!\in\! \{0,1, \ldots, T\},l \!\in\! \{0,1, \ldots, L\}\}$ to the set of all clients’ model parameters and gradients over time:
\begin{align} \label{mapping}
&\small \Gamma(\{\boldsymbol{w_{i,l}^{t}}, \!\nabla \! F_{i} (\boldsymbol{w_{i,l}^{t}}) \!\mid\! t \!\in\! \{0,1, \ldots, T\}, l \!\in\! \{0,1, \ldots, L\}\}) \nonumber \\
\!\!\!=\! (&\small \Gamma_{0}^{0}(\{\boldsymbol{w_{i,l}^{t}},\! \nabla \! F_{i} (\boldsymbol{w_{i,l}^{t}}) \!\mid\! t \!\in\! \{0,1, \ldots, T\}, l \!\in\! \{0,1, \ldots, L\}\}), \ldots, \nonumber \\
&\small \Gamma_{L}^{0}(\{\boldsymbol{w_{i,l}^{t}},\! \nabla \! F_{i} (\boldsymbol{w_{i,l}^{t}}) \!\mid\! t \!\in\! \{0,1, \ldots, T\}, l \!\in\! \{0,1,\ldots, L\}\}), \ldots, \nonumber \\
&\small \Gamma_{0}^{T}(\{\boldsymbol{w_{i,l}^{t}}, \! \nabla \! F_{i} (\boldsymbol{w_{i,l}^{t}}) \!\mid\! t \!\in\! \{0,1, \ldots, T\}, l \!\in\! \{0,1, \ldots, L\}\}), \ldots, \nonumber \\
&\small \Gamma_{L}^{T}(\{\boldsymbol{w_{i,l}^{t}}, \!\! \nabla \! F_{i} (\boldsymbol{w_{i,l}^{t}}) \!\mid\! t \!\in\! \{0,1, \ldots,  T\}, l \!\in\! \{0,1, \ldots, L\}\})). \!\!\!
\end{align}  


Thus, the fixed point of mapping $\Gamma (\{\boldsymbol{w_{i,l}^{t}}, \nabla F_{i} (\boldsymbol{w_{i,l}^{t}}) \mid t \in \\ \{0,1, \ldots, T\}, l \!\in \{0, 1, \ldots, L\}\}) = \Gamma (\{\boldsymbol{w_{i,l}^{t}}, \nabla  F_{i} (\boldsymbol{w_{i,l}^{t}}) \mid  t \!\in\! \\ \{0, 1, \ldots, T\},  l \!\in\! \{0,1, \ldots, L\}\})$ in Eq.~(\ref{mapping}) should be reached by enforcing $\boldsymbol{\phi_{1,l}^{t}}$ and $\boldsymbol{\phi_{2,l}^{t}}$ to replicate $\frac{1}{N}\!\!\sum_{i=1}^{N}\!\! \nabla{F_{i} (\boldsymbol{w_{i,l}^{t})}}$ and $\frac{1}{N}\!\! \sum_{i=1}^{N} \! \boldsymbol{w_{i,l}^{t}}$ respectively. Then, based on Assumption~\ref{bounded_Gradients} and Assumption~\ref{assumption_bounded_parameter}, we define a continuous space $ \Omega \!=\! [-Q, Q]_{0}^{0} \times \ldots \times[-Q, Q]_{L}^{0} \times \ldots \times[-Q, Q]_{0}^{T}\times \ldots \times[-Q, Q]_{L}^{T} \times [-P, P]_{0}^{0} \times \ldots \times[-P, P]_{L}^{0} \times \ldots \times[-P, P]_{0}^{T}\times \ldots \times[-P, P]_{L}^{T}$ for $ \boldsymbol{w_{i,l}^{t}}$ and $ \nabla{F_{i} (\boldsymbol{w_{i,l}^{t}})}$ in $L \times T$ dimensions. Since each $ \Gamma_{l}^{t}$ is continuous in the space $ \Omega$, $ \Gamma$ is similarly a continuous mapping from $ \Omega$ to $ \Omega$. Thus, according to Brouwer's fixed-point theorem, the fixed point of mean-field terms exists. \end{proof}

The existence of a fixed point can not inherently guarantee that the mapping will converge to that fixed point. Consequently, we introduce the following theorem to demonstrate the fixed point's convergence in the feasible domain.
\begin{proposition} \label{proposition_fixed_point_reach}
The fixed points of mean-field terms $\boldsymbol{\phi_{1,l}^{t}}$ and $\boldsymbol{\phi_{2,l}^{t}}$ can be attained through an iterative algorithm.
\end{proposition}

The proof of Proposition \ref{proposition_fixed_point_reach} is presented in Appendix~B of the supplementary file. We summarize the pretraining process of iteratively figuring out the fixed point of mean-field terms $\boldsymbol{\phi_{1,l}^{t}}$ and $ \boldsymbol{\phi_{2,l}^{t}}$ in Algorithm \ref{alg1}, which attains linear computation complexity $O(TNLJ)$, where $J$ represents the number of iterative rounds required to achieve the fixed point. Given global model $ \boldsymbol{w_{i,0}^{t}}$, initial value for $\boldsymbol{\phi_{1,l}^{t}}(0)$, $\boldsymbol{\phi_{2,l}^{t}}(0)$ and $\eta_{i,l}^{t}$, we can easily obtain $\eta_{i,l}^{t}$ according to Theorem \ref{theorem_optimal_lr}. Then, based on $\eta_{i,l}^{t}$ and $ \boldsymbol{w_{i,0}^{t}}$, we can calculate $ \boldsymbol{w_{i,l}^{t}}$ and $ \nabla{F_{i}(\boldsymbol{w_{i,l}^{t})}}$ by Eq.~(\ref{local_model_update_mean_field}). Accordingly, mean-field terms $ \boldsymbol{\phi_{1,l}^{t}}$ and $ \boldsymbol{\phi_{2,l}^{t}}$ can be updated. Repeating the process until $\boldsymbol{\phi_{1,l}^{t,\texttt{est}}}(j + 1)\! \rightarrow \! \boldsymbol{\phi_{1,l}^{t,\texttt{est}}}(j)$ and $\boldsymbol{\phi_{2,l}^{t,\texttt{est}}}(j + 1)\! \rightarrow \! \boldsymbol{\phi_{2,l}^{t,\texttt{est}}}(j)$ within preset error $ \epsilon$ simultaneously, the fixed point is subsequently found. In the meanwhile, $ \eta_{i,l}^{t}$, $ i \!\in\! \{1,  2, \ldots, N\}$ in each instant is obtained. The workflow of our proposed FedAgg is showcased in Algorithm~\ref{alg2}. 

\begin{algorithm}[t]
\caption{Adaptive Federated Learning with the Aggregated Gradients (FedAgg)} 
\begin{algorithmic}[1]
\State {\bf Input: } 
$N$, $L$, $T$, weight $\alpha$, initial global model $\boldsymbol{w_{i,0}^{t}}$, mean-field terms $\boldsymbol{\phi_{1,l}^{t}}$ and $\boldsymbol{\phi_{2,l}^{t}}$, $ t \!\in\! \{0, 1, \ldots, T\},l \!\in\! \{0,\newline 1,  \ldots, L\}, i \!\in\! \{1, 2, \ldots, N\}$.
\For{$t=0$ to $T$}
    \State Calculate adaptive learning rate $\eta_{i,l}^{t}$ based on Eq.~(\ref{optimal_learning_rate})
    \For{each client $i \in\{0,1, \ldots, N\}$}
        \For{$l=0$ to $L-1$}
            \State Local training: $\boldsymbol{w_{i,l+1}^{t}}\!=\!\boldsymbol{w_{i,l}^{t}}\!-\!\eta_{i,l}^{t} \nabla{F_{i}(\boldsymbol{w_{i,l}^{t})}}$
        \EndFor
    \EndFor
\State Update global model: $\boldsymbol{\bar w^{t+1}} = \sum_{i=1}^{N} \frac{D_{i}}{\sum_{j=1}^{N} D_{j}} \boldsymbol{w_{i,L}^{t}}$
\EndFor
\State \Return The optimal global model parameter $\boldsymbol{\bar w^{T}}$.
\end{algorithmic}
\label{alg2}
\end{algorithm}

\begin{remark}
Before the formal FL training, the central server precomputes the mean-field terms with linear computation complexity and transmits them to the participants along with the initial global model $\boldsymbol{\bar w^{0}}$. Thus, the mean-field terms can be regarded as known functions for the clients. Moreover, the model information required for mean-field terms calculation can be obtained from the historical training process, thereby mitigating local information exchange and reducing the risk of privacy leakage during the formal FL training.
\end{remark}

\section{Convergence analysis}\label{Convergence_analysis}
In this section, we conduct comprehensive theoretical analyses regarding the convergence of our proposed FedAgg, thereby establishing a robust theoretical foundation. At the inception of iteration $t$, we suppose that all clients are conversant with the global descent direction, and our descent direction analysis is executed at $ \boldsymbol{w_{i,0}^{t}} \! = \! \boldsymbol{\bar w^{t}}$. Firstly, we probe into the upper bound of the learning rate $ \eta_{i,l}^{t}$. From Proposition~\ref{proposition_lr_linear_combination}, $\eta_{i,l}^{t}$ is the linear combination of $ \boldsymbol{w_{i,l}^{t}}$ and mean-field terms, based on which, we derive a more stringent upper bound for $ \eta_{i,l}^{t}$ in the following proposition.
\begin{proposition} \label{proposition_learning_rate_bound}
(Bounded Learning Rate) For any $ \eta_{i,l}^{t}$ in training epochs, there exists a tight upper bound $ \delta_{i,l}^{t}$ such that $ \eta_{i,l}^{t} \!\in\! (0, \delta_{i,l}^{t}]$ with $ \delta_{i,l}^{t} \!<\! 1$.
\end{proposition}

The proof of Proposition \ref{proposition_learning_rate_bound} is delineated in Appendix~C of the supplementary file. It has been proved in \cite{karimireddy2020scaffold} that training on the heterogeneous dataset (Non-IID dataset) can impede the model's convergence performance and substantially decelerate the convergence rate. Thus, we rigorously establish an upper bound for the client drifting term $ \|\boldsymbol{w_{i,l}^{t}} - \boldsymbol{\bar w^{t}}\|$, which significantly influences the efficacy of the global model and verifies the robustness of FedAgg under the Non-IID scenarios.
\begin{proposition} \label{proposition_client_drifting}
(Bounded Client Drifting Term) Suppose~$\delta_{l}\!=\!\text{min}\{\eta_{i,l}^{t}\}$, $ \delta_{h}\!=\!\text{max}\{\delta_{i,l}^{t}\}$, and $F_{i}(\boldsymbol{w_{i,l}^{t}})$ satisfies Assumption~\ref{assumption_smoothness}, for each client $i$, $\|\boldsymbol{w_{i,l}^{t}} - \boldsymbol{\bar w^{t}}\| \leq L P \delta_{h}$ holds. 
\end{proposition}

\begin{figure*}[t]
\setlength{\abovecaptionskip}{2pt} 
    \centering
    \subfloat[IID]{
        \label{cifar100-IID}
        \includegraphics[width=0.32\textwidth, trim=50 20 140 40,clip]{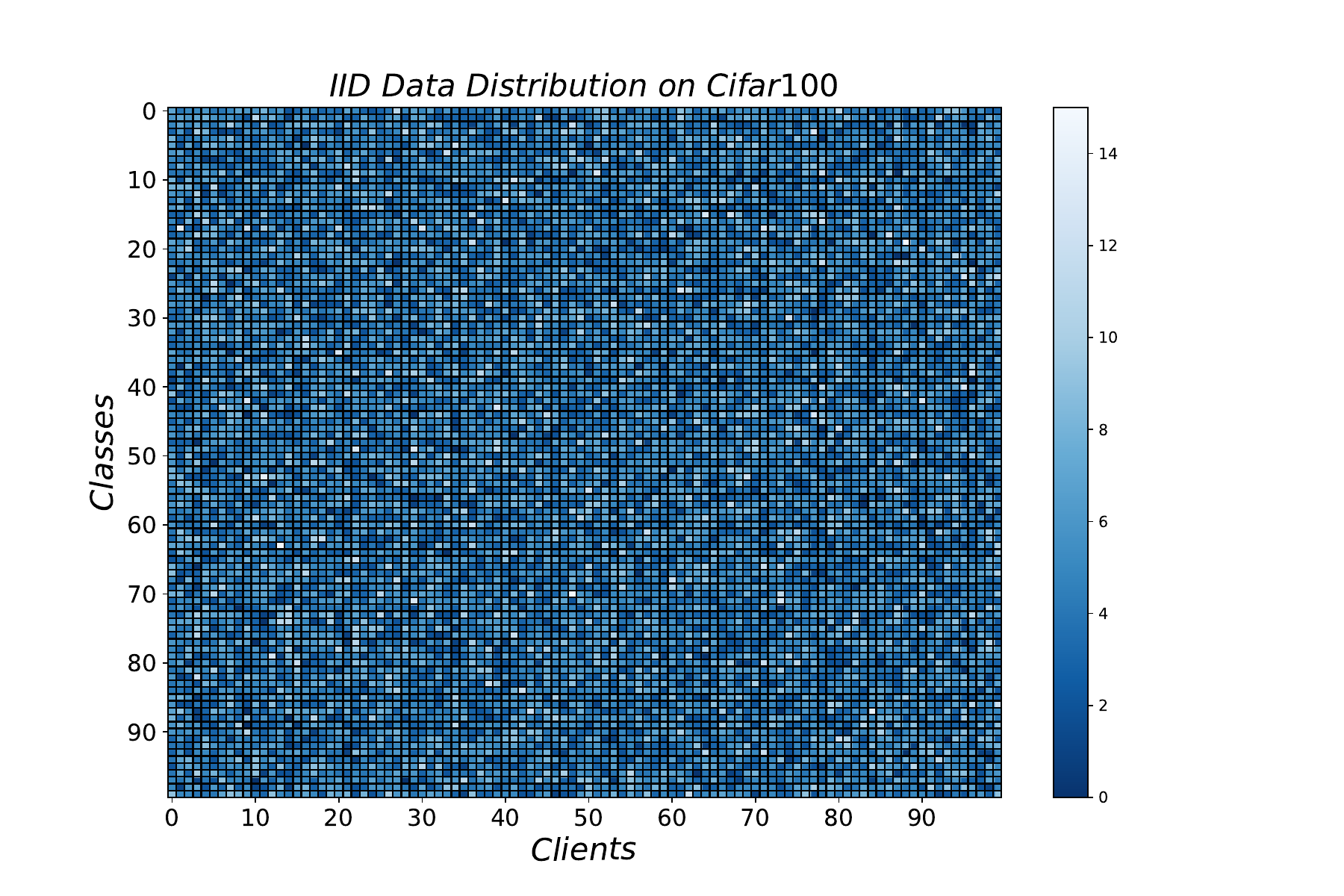}}
    \subfloat[Dir-0.6]{
        \label{cifar100-dir06}
        \includegraphics[width=0.32\textwidth, trim=50 20 140 40,clip]{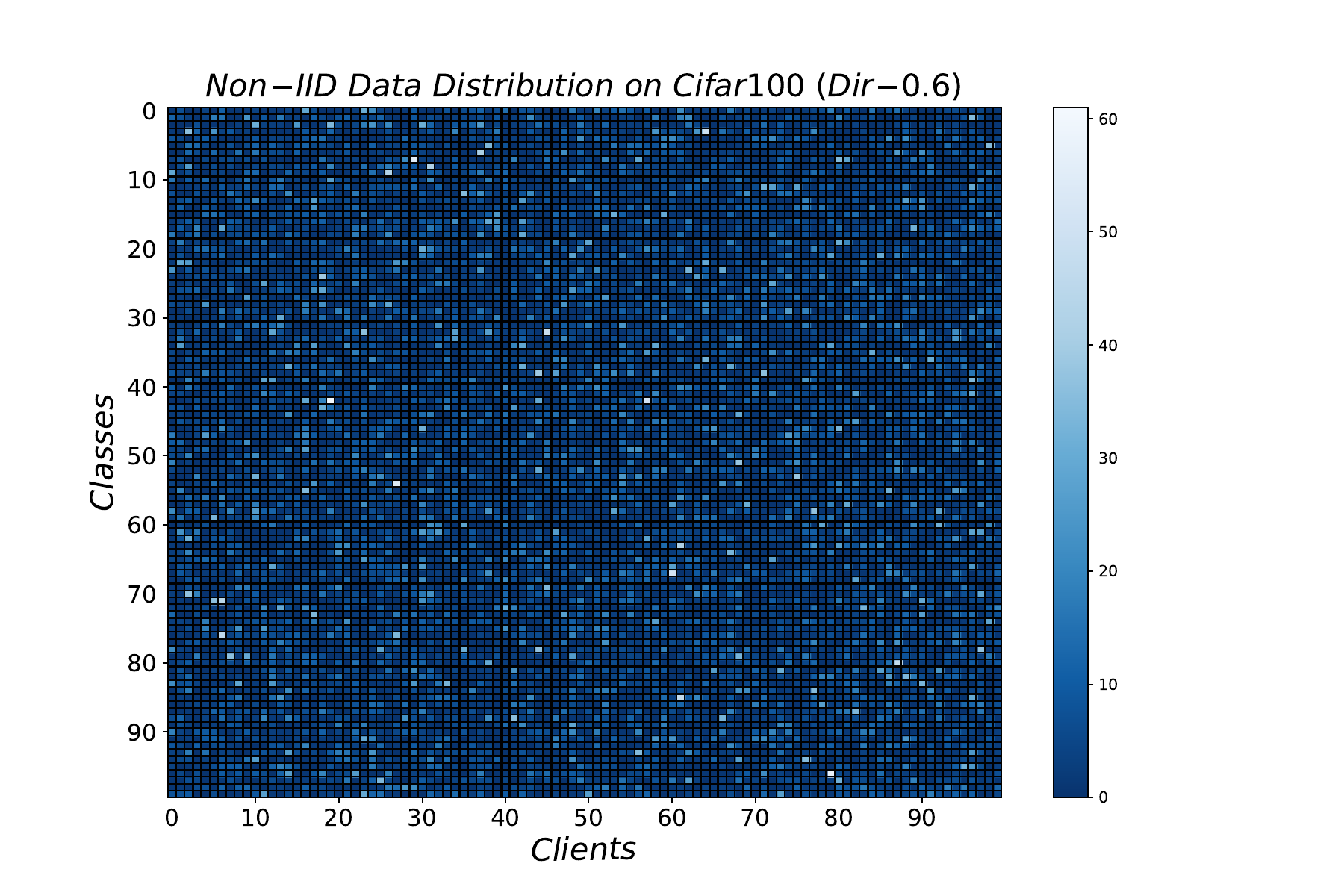}}
    \subfloat[Wasserstein Distance]{
        \label{cifar100-emd}
        \includegraphics[width=0.35\textwidth, trim=20 10 50 20,clip]{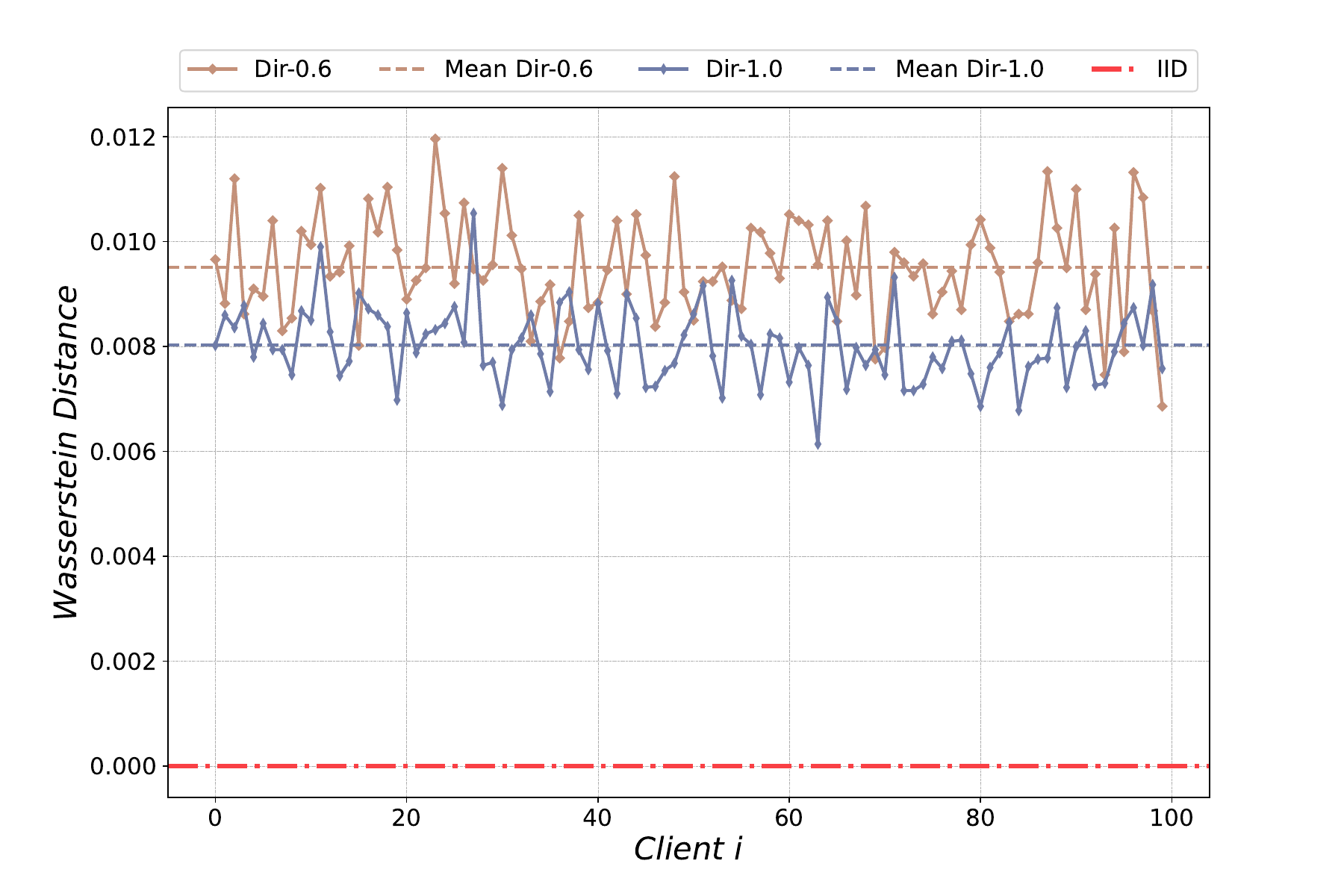}}
\caption{The data distribution on CIFAR-100 dataset}
\label{dirichlet distribution on cifar100}
\vspace{-10pt}
\end{figure*}

The proof of Proposition \ref{proposition_client_drifting} is presented in Appendix~D of the supplementary file. The optimality gap $ \|\boldsymbol{\bar w^{T}} - \boldsymbol{w^{*}}\|_{2}^{2}$ serves as a critical metric for the global model accuracy assessment \cite{ding2023incentive}, thus we derive an upper bound for this gap within the subsequent theorem, considering the aggregated gradients.
\begin{theorem} \label{theorem_model_gap}
Suppose the global loss function $ F(\boldsymbol{w})$ satisfies Assumption~\ref{assumption_smoothness}. If the global loss function satisfies $\psi$-strong convex in Assumption \ref{assumption_convexity}, after $T$ iteration of FL training, the expected optimality gap of the global model is given by
\begin{align} \label{model_gap_convex}
\!\! \mathbb{E}[\|\boldsymbol{\bar w^{T}} \!\!-\! \boldsymbol{w^{*}}\|_{2}^{2}] \!\leq&\ \sum\nolimits_{k=0}^{T-1} (1 \!-\! \frac{\delta_{h} L \psi}{N})^{k} P^{2} \! L^{2} \! \delta_{h}^{2} (\beta L \delta_{h}\!+\! 1)  \nonumber \\
& + (1 \!-\! \frac{\delta_{h} L \psi}{N})^{T} \mathbb{E}[\|\boldsymbol{\bar w^{0}} \!-\!  \boldsymbol{w^{*}}\|_{2}^{2}].
\end{align}
Moreover, for the non-convex global loss function, the expected optimality gap of the global model is upper bounded~by:
\begin{align} \label{model_gap_nonconvex}
\mathbb{E}[\|\boldsymbol{\bar w^{T}} \!\!-\! \boldsymbol{w^{*}}\|_{2}^{2}] \!\leq& \frac{T (T \!-\! 1) P \delta_{h}^{2} L^{2} \|g^\top\|_{2}}{N} \!+\! \text{\small $T P^{2} \! L^{2} \! \delta_{h}^{2} (\beta L \delta_{h}\!+\! 1) $} \nonumber \\
&+\!\! \left(\!1 \!+\! \frac{2 \delta_{h} L T \|g^\top\|_{2}}{N} \! \right)\! \mathbb{E}[\|\boldsymbol{\bar w^{0}} \!\!-\! \boldsymbol{w^{*}}\|^{2}_{2}]. \!\!
\end{align}
\end{theorem}

The proof of Theorem \ref{theorem_model_gap} is provided in Appendix~E of the supplementary file. According to Propositions~\ref{proposition_learning_rate_bound}-\ref{proposition_client_drifting}, we proceed to formulate a convergence upper bound for the consecutive change of the FL global loss function $F(\boldsymbol{\bar w^{t}})$.
\begin{proposition} \label{proposition_difference}
(One Round Progress) With the model aggregation weight $\theta_{i} \!=\! \frac{D_{i}}{\sum\nolimits_{j=1}^{N} D_{j}}$, for any two consecutive global training iterations $t$ and $t+1$, the convergent upper bound of global loss function satisfies:
\begin{align} \label{difference}
& F(\boldsymbol{\bar w^{t+1}}) - F(\boldsymbol{\bar w^{t}}) \nonumber \\
\leq & \frac{\beta\delta_{l}}{N}  \sum\nolimits_{i=1}^{N}\!\sum\nolimits_{k=0}^{L-1}\!\sum\nolimits_{j=1}^{N}\!\theta_{i}\|\boldsymbol{w_{j,k}^{t}}\!-\!\boldsymbol{\bar w^{t}}\|_{2}\|\nabla \! {F(\boldsymbol{\bar w^{t}})}\|_{2}    \nonumber \\
&+ \beta^{2} L \delta_{h}^{2}\sum\nolimits_{i=1}^{N} \sum\nolimits_{k=0}^{L-1}\sum\nolimits_{j=1}^{N} \theta_{i}^{2}\|\boldsymbol{w_{j,k}^{t}}\!-\!\boldsymbol{\bar w^{t}}\|_{2}^{2}  \nonumber \\
& + (\beta L^{2} \delta_{h}^{2} -L\delta_{l})  \|\nabla \! {F(\boldsymbol{\bar w^{t}})}\|_{2}^{2} \nonumber \\
\leq& (\beta L^{2} \delta_{h}^{2} \!-\! L \delta_{l})\|\nabla \! F(\boldsymbol{\bar w^{t}})\|_{2}^{2}\!+\! L^{2}\! P^{2}\! \beta \delta_{l}\delta_{h} \!+\! \beta^{2}\! L^{4}\! N\! P^{2}\! \delta_{h}^{4}. \!
\end{align}
\end{proposition}

The detailed proof of Proposition \ref{proposition_difference} is elaborated in Appendix~F of the supplementary file. As delineated by Eq.~(\ref{difference}), our derivation reveals that the one-round convergence bound for our proposed adaptive FL algorithm FedAgg intrinsically pertains to the norm of the gradient of the global parameter at iteration $t$ and predetermined constants, ensuring the existence of a trustworthy convergence upper bound. Building upon the foundation of Proposition \ref{proposition_difference}, we further extend our analysis on the upper bound of the convergence rate and convergence error in the following theorems.


\begin{theorem} \label{theorem_convergence_rate}
(Convergence Rate) After $T$ iterations of FL model training, the expected reduction of global loss function is upper bounded by:
\begin{align} \label{convergence_rate}
&\mathbb{E} [F(\boldsymbol{\bar w^{T}})] \!-\! F(\boldsymbol{w^{*}}) \nonumber \\
\leq&  \sum\nolimits_{k=1}^{T-1} \! (1 \!+\! 2 \mu (\beta L^{2}\! \delta_{h}^{2} \!-\! L \delta_{l}))^{k} (L^{2}\! P^{2}\! \beta \delta_{l}\delta_{h} \!+\! \beta^{2}\! L^{4}\! N\! P^{2}\! \delta_{h}^{4})  \nonumber \\
& + (1 \!+\! 2 \mu (\beta L^{2} \delta_{h}^{2} \!-\! L \delta_{l}))^{T} \! (\mathbb{E}[F(\boldsymbol{\bar w^{0}})] \!-\! F(\boldsymbol{w^{*}})).
\end{align}
\end{theorem}

\begin{theorem} \label{theorem_convergence_error}
(Convergence Error) The average of expected squared $\ell_{2}$-norm of the gradient is upper bounded by:
\begin{align} \label{convergence_error}
\!\!\! \frac{1}{T}\! \sum\nolimits_{t=0}^{T-1}\! \mathbb{E}[ \|\nabla \! F(\boldsymbol{\bar w^{t}})\|_{2}^{2}] \leq&\  \frac{F(\boldsymbol{\bar w^{0}}) - F(\boldsymbol{w^{*}})}{T (L \delta_{l} -  \beta L^{2} \delta_{h}^{2})} \nonumber \\
&+ \! \frac{P^{2} L\beta\delta_{h} (\delta_{l} \!+\! \beta L^{2} N \delta_{h}^{3})}{\delta_{l} - \beta L \delta_{h}^{2}} . 
\end{align}
\end{theorem}

The detailed proofs of Theorem~\ref{theorem_convergence_rate} and Theorem~\ref{theorem_convergence_error} refer to Appendix~G-H of the supplementary file. Moreover, delving into the foundational assumptions for our above convergence analyses, we provide a pertinent remark concerning the characteristics of the global loss function, specific to its convex and non-convex feasibility.
\begin{remark}
(Convex \& Non-Convex Applicability) From the above convergence analyses, the evaluation of FedAgg exclusively relies on the smoothness assumption, which indicates that the effectiveness of our results transcends the convexity paradigm of the function in question. Specifically, we conduct convergence analyses of FedAgg across both convex and non-convex domains, which broadens the applicability of our algorithm for addressing distributed optimization challenges.
\end{remark}

\section{Numerical Experiments}\label{Numerical_Experiments}
In this section, we conduct extensive experiments on the MNIST, CIFAR10, EMNIST-L, and CIFAR100 datasets to verify the performance of our proposed adaptive learning FL algorithm FedAgg. We first introduce the experiment setup, including the experiment platform, datasets, data partition, local training model, baseline FL methods, and hyperparameter settings. Then, we compare the performance of FedAgg with other FL algorithms in terms of data distributions, client participation ratio, adaptive strategies, different local model architecture, and hyperparameter efficiency. Experimental results show that our method outperforms the benchmarks.

\begin{table}[t]
\setlength{\abovecaptionskip}{-1pt} 
\renewcommand\arraystretch{1.0}
\caption{Basic Information of Datasets}
\begin{center}
\resizebox{0.48\textwidth}{!}{\begin{tabular}{c|c|c|c|c}
\toprule[1pt]
\textbf{Datasets}&\textbf{Training Set Size}&\textbf{Test Set Size}&\textbf{Class} &\textbf{Image Size}  \\ \cmidrule[0.5pt](l{1pt}r{0pt}){1-5}

MNIST & 60,000 & 10,000  & 10 & 1 $\times$ 28 $\times$ 28  \\ \cmidrule[0.5pt](l{1pt}r{0pt}){1-5}

CIFAR-10 & 50,000 & 10,000  & 10 & 3 $\times$ 32 $\times$ 32  \\ \cmidrule[0.5pt](l{1pt}r{0pt}){1-5}

EMNIST-L & 124,800 & 20,800  & 26 & 1 $\times$ 28 $\times$ 28  \\ \cmidrule[0.5pt](l{1pt}r{0pt}){1-5}

CIFAR-100 & 50,000 & 10,000  & 100 & 3 $\times$ 32 $\times$ 32  \\ 
\bottomrule[1pt]
\end{tabular}}
\label{basic_information_of_datasets}
\end{center}
\vspace{-12pt}
\end{table}

\subsection{Experiment Setups}

\subsubsection{Platform Setup}
Experiments are conducted on a workstation (CPU: Intel(R) Xeon(R) Gold 6133 @ 2.50GHz; RAM: 32GB DDR4 2666 MHz; GPU: NVIDIA GeForce RTX 4090 with CUDA version 12.2). The simulator is composed of three parts: (\romannumeral 1) The data partitioning part that partitions the datasets as IID and Non-IID distribution. (\romannumeral 2) The model training part instantiates heterogeneous FL clients for local model training, aggregates, and updates the global model until reaching pre-fixed global training iteration or threshold. (\romannumeral 3) The estimator calculating part finds the fixed point for mean-field terms $\boldsymbol{\phi_{1,l}^{t}}$ and $\boldsymbol{\phi_{2,l}^{t}}$. We utilize the PyTorch library \cite{paszke2019pytorch} and the environment is built on Python 3.11.

\begin{table*}[t]
\centering
\setlength{\abovecaptionskip}{3pt} 
\renewcommand\arraystretch{1.0}
\caption{Accuracy on different datasets with 20\% and 100\% participant ratio.}
\resizebox{1.0\textwidth}{!}{\begin{tabular}
{c|c|cc|cc|cc|cc|cc|cc|cc} 
\toprule[1.2pt]
\multirow{2}{*}{\multirowcell{2}{\centering\textbf{Datasets}}} & \multirow{2}{*}{\multirowcell{2}{\centering\textbf{Distribution}}} & \multicolumn{2}{c|}{\parbox{1.5cm}{\centering\textbf{FedAvg}\cite{mcmahan2017communication}}} & \multicolumn{2}{c|}{\parbox{1.8cm}{\centering\textbf{FedAdam}\cite{reddi2021adaptive}}} & \multicolumn{2}{c|}{\parbox{1.6cm}{\centering\textbf{FedProx}\cite{li2020federated}}} & \multicolumn{2}{c|}{\parbox{1.6cm}{\centering\textbf{FedDyn}\cite{durmus2021federated}}} & \multicolumn{2}{c|}{\parbox{1.8cm}{\centering\textbf{FedBABU}\cite{oh2022fedbabu}}} & \multicolumn{2}{c|}{\parbox{1.5cm}{\centering\textbf{FedGH}\cite{yi2023fedgh}}} & \multicolumn{2}{c}{\parbox{1.5cm}{\centering\textbf{FedAgg}(Ours)}}    \\ \cmidrule[0.5pt](l{1pt}r{0pt}){3-16}

&  & 20\% & 100\% & 20\% & 100\% & 20\% & 100\% & 20\% & 100\% & 20\% & 100\% & 20\% & 100\% & 20\% & 100\% \\ \cmidrule[0.8pt](l{1pt}r{0pt}){1-16}

\multirow{2}{*}{\multirowcell{2}{\parbox{2.1cm}{\centering \textbf{MNIST} \\ \textbf{+Linear}}}}
        & IID & 88.23 & 88.24 & 89.85 & 89.99 & 88.26 & 88.29 & 88.28 & 88.30 & - & - & 90.21 & 90.29 & \textbf{90.35} & \underline{90.77}  \\ \cmidrule[0.5pt](l{1pt}r{0pt}){2-16}
        
        & Non-IID & 86.28 & 86.57 & 89.08 & \underline{89.60} & 86.19 & 86.63 & 85.18 & 86.62 & - & - & 89.02 & 88.97 & \textbf{89.45} & 89.56 \\ \midrule[0.8pt]

\multirow{2}{*}{\multirowcell{2}{\parbox{2.1cm}{\centering \textbf{MNIST} \\ \textbf{+MNIST-CNN}}}}
        & IID & 93.59 & 93.94 & 96.55 & 96.59 & 93.35 & 93.62 & 93.56 & 94.06 & 96.96 & 97.26 & 97.21 & 97.42 & \textbf{97.48} & \underline{97.52} \\ \cmidrule[0.5pt](l{1pt}r{0pt}){2-16}
        
        & Non-IID & 86.93 & 85.35 & 87.59 & 90.70 & 85.96 & 87.30 & 84.48 & 86.35 & 90.84 & 93.29 & 94.06 & \underline{94.67} & \textbf{94.91} & 93.51 \\ \midrule[0.8pt]

\multirow{3}{*}{\multirowcell{3}{\textbf{CIFAR-10} \\ \textbf{+CIFAR-10-CNN}}}
        & IID & 66.97 & 67.44 & 68.22 & 69.93 & 65.31 & 66.82 & 65.89 & 66.47 & 67.55 & 67.96 & 68.38 & 70.14 & \textbf{69.21} & \underline{70.48} \\ \cmidrule[0.5pt](l{1pt}r{0pt}){2-16}
        
        & Dir-0.6 & 64.95 & 64.53 & 64.92 & 65.58 & 63.76 & 65.98 & 65.21 & 66.03 & 66.01 & 67.27 & 66.77 & 67.12 & \textbf{66.96} & \underline{67.30} \\ \cmidrule[0.5pt](l{1pt}r{0pt}){2-16}
        
        &  Dir-1.0 & 65.69 & 65.87 & 67.85 & 67.88 & 65.49 & 66.38 & 64.20 & 65.90 & 66.67 & 67.39 & 67.79 & 67.86 & \textbf{68.06} & \underline{68.46} \\ \midrule[0.8pt]

\multirow{3}{*}{\multirowcell{3}{\textbf{EMNIST-L} \\ \textbf{+LeNet-5}}}
        & IID & 92.52 & 93.14 & 93.09 & 92.79 & 92.77 & 92.79 & 92.52 & 92.80 & 93.42 & 93.40 & 93.39 & 93.72 & \textbf{94.11} & \underline{94.39} \\ \cmidrule[0.5pt](l{1pt}r{0pt}){2-16}
        
        & Dir-0.6 & 90.98 & 91.13 & 92.60 & 92.73 & 91.84 & 91.93 & 91.61 & 92.38 & 92.95 & 93.25 & 92.24 & 93.25 & \textbf{93.67} & \underline{93.68} \\ \cmidrule[0.5pt](l{1pt}r{0pt}){2-16}
        
        &  Dir-1.0 & 92.18 & 92.75 & 92.89 & 92.74 & 92.44 & 92.58 & 92.25 & 92.55 & 93.07 & 93.28 & 93.19 & 93.53 & \textbf{94.10} & \underline{93.88} \\ \midrule[0.8pt]

\multirow{3}{*}{\multirowcell{3}{\textbf{CIFAR-100} \\ \textbf{+VGG-11}}}
        & IID & 44.77 & 44.46 & 46.58 & 45.79 & 45.08 & 44.06 & 45.30 & 44.23 & 50.06 & 51.08 & 47.52 & 48.24 & \textbf{50.91} & \underline{51.22} \\ \cmidrule[0.5pt](l{1pt}r{0pt}){2-16}
        
        & Dir-0.6 & 44.45 & 44.44 & 47.48 & 47.43 & 44.96 & 45.25 & 45.23 & 44.93 & 48.79 & 50.02 & 46.06 & 47.13 & \textbf{49.98} & \underline{50.54} \\ \cmidrule[0.5pt](l{1pt}r{0pt}){2-16}
        
        &  Dir-1.0 & 45.04 & 45.08 & 47.85 & 46.39 & 45.53 & 45.49 & 45.26 & 45.09 & 49.77 & 50.55 & 47.09 & 47.95 & \textbf{50.59} & \underline{50.73} \\ 
\bottomrule[1.2pt]
\end{tabular}}
\label{accuracy}
\end{table*}

\begin{figure*}[t]
\setlength{\abovecaptionskip}{3pt} 
\centerline{\includegraphics[width=1.0\textwidth, trim=50 60 50 70,clip]{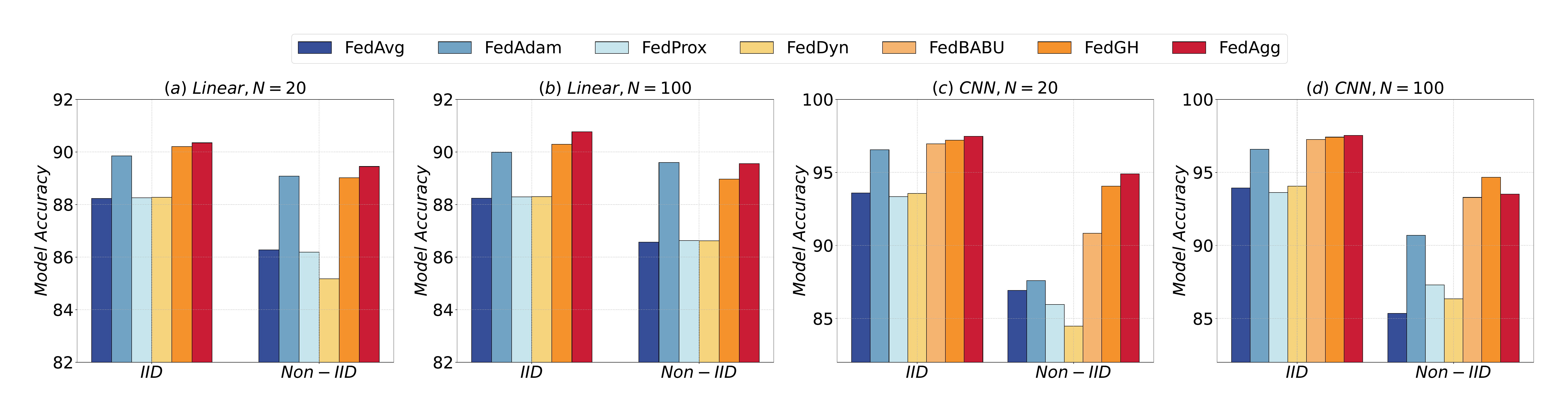}}
\caption{Accuracy on MNIST dataset with 20\% and 100\% of participating clients different data heterogeneity.}
\label{MNIST_barplot}
\vspace{-8pt}
\end{figure*}

\begin{figure*}[t]
\setlength{\abovecaptionskip}{3pt} 
\centerline{\includegraphics[width=1.0\textwidth, trim=0 10 0 5,clip]{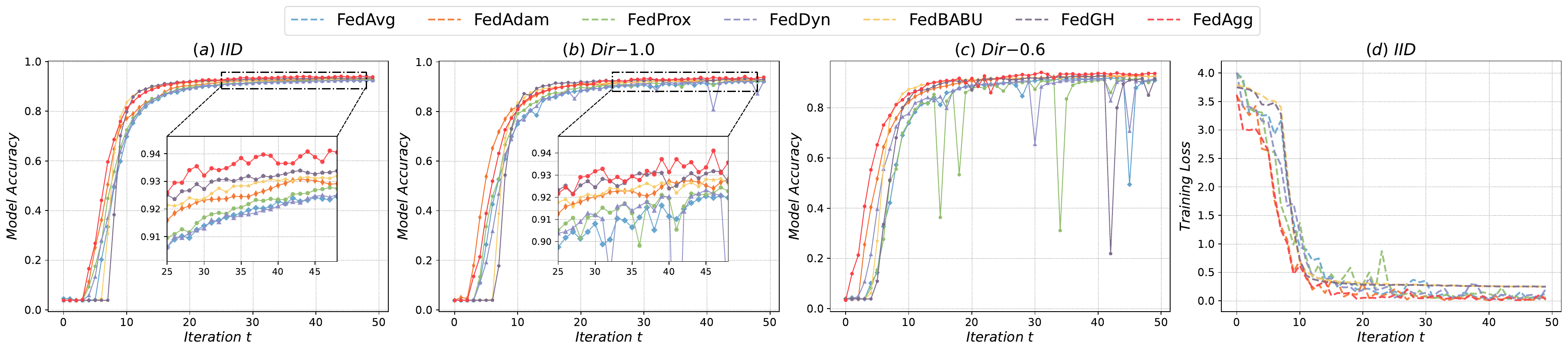}}
\caption{Accuracy and loss on EMNIST-L dataset with 20\% of participating clients across different levels of data heterogeneity.}
\label{EMNIST_N20}
\vspace{-12pt}
\end{figure*}

\begin{figure*}[t]
\setlength{\abovecaptionskip}{3pt} 
\centerline{\includegraphics[width=1.0\textwidth, trim=0 10 0 5,clip]{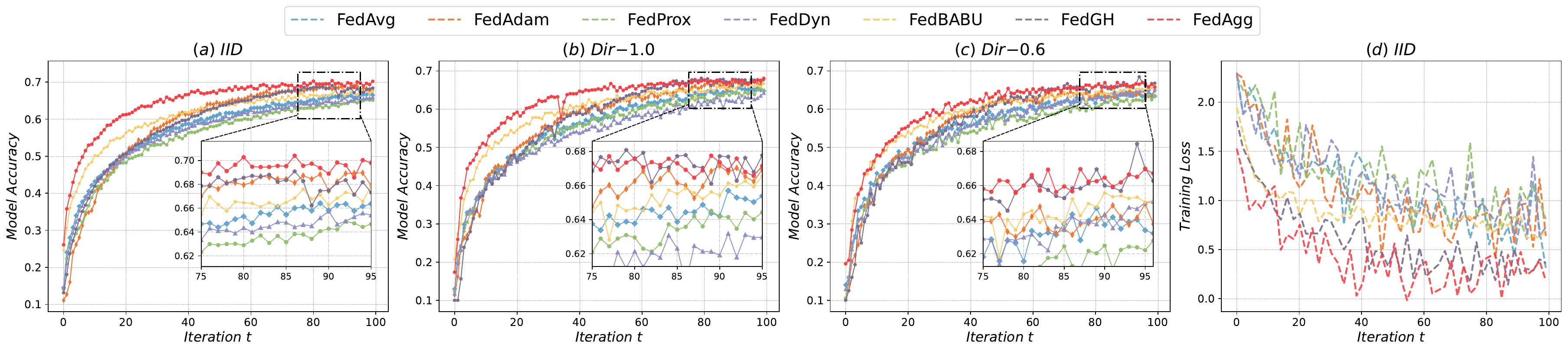}}
\caption{Accuracy and loss on CIFAR-10 dataset with 20\% of participating clients across different levels of data heterogeneity.}
\label{Cifar10_N20}
\vspace{-12pt}
\end{figure*}

\subsubsection{Datasets} 
We conduct our experiments on four typical and real-world datasets with different types and sorts (i.e. MNIST, EMNIST-L, CIFAR-10, and CIFAR-100) to validate the performance of our proposed algorithm FedAgg. Specifically, (\romannumeral 1) MNIST dataset contains 70,000 gray-scale hand-written digit images in total which are divided into 60,000 images as a training set and 10,000 images as a test set. Besides, the images in the MNIST dataset are 28 $\times$ 28 pixels, with a total of 10 classes. (\romannumeral 2) EMNIST-L dataset contains 145,600 gray-scale hand-written letter images in total, which are divided into 124,800 images as a training set and 20,800 images as a test set. Furthermore, the EMNIST-L dataset comprises gray-scale hand-written letter images measuring 28 $\times$ 28 pixels, encompassing a total of 26 classes. (\romannumeral 3) CIFAR-10 and CIFAR-100 datasets are both composed of 60,000 RGB images, which are divided into 50,000 images as a training set and 10,000 images as a test set with images 32 $\times$ 32 pixels. For the CIFAR-10 dataset, 10 classes of images are contained, whereas the CIFAR-100 dataset consists of 100 classes, posing a significant challenge for model training. The basic information of the datasets is summarized in Table~\ref{basic_information_of_datasets}.

\begin{figure*}[t]
\setlength{\abovecaptionskip}{3pt} 
\centerline{\includegraphics[width=1.0\textwidth, trim=0 10 0 5,clip]{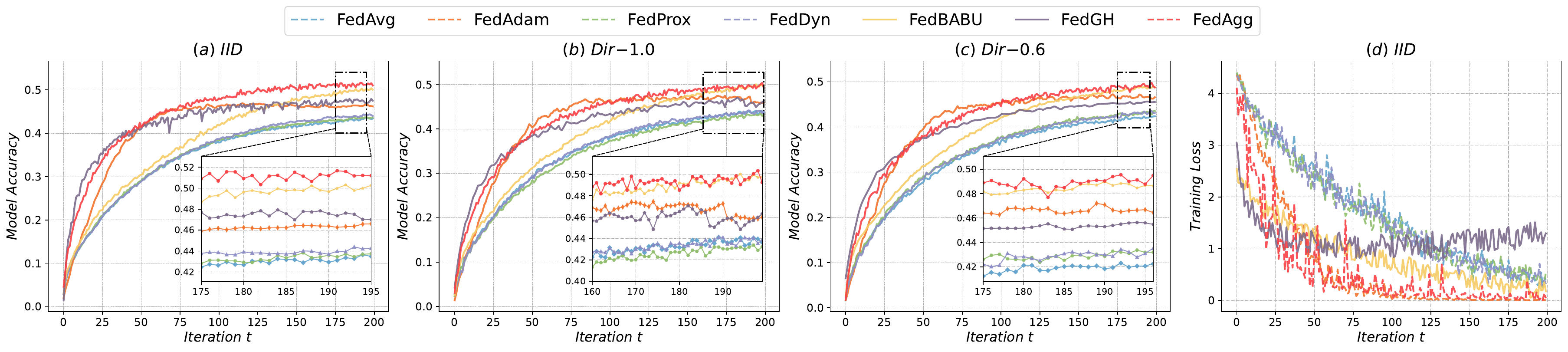}}
\caption{Accuracy and loss on CIFAR-100 dataset with 20\% of participating clients across different levels of data heterogeneity.}
\label{Cifar100_N20}
\vspace{-8pt}
\end{figure*}

\subsubsection{Data Partition}
We adopt two different data partition methods for datasets. (\romannumeral 1) Similar to \cite{mcmahan2017communication}, we use the pathological Non-IID partition method for the MNIST dataset. (\romannumeral 2) For the rest datasets, we utilize Dirichlet distribution \cite{hsu2019measuring} to partition datasets as Non-IID settings and vary the Dirichlet parameter $\sigma \! \in \! \{0.6, 1.0, \infty\}$ to validation the robustness of our proposed algorithm FedAgg. In Fig.~\ref{dirichlet distribution on cifar100}, we display the heat map of the data distribution on CIFAR-100, where Fig.~\ref{dirichlet distribution on cifar100}\subref{cifar100-IID} and Fig.~\ref{dirichlet distribution on cifar100}\subref{cifar100-dir06} represent the data distribution with IID setting and with Dirichlet parameter $\sigma \! = \! 0.6$ respectively. The color of the heat map is shallow under the IID distribution with a close amount of labels. For Non-IID distribution, the heat map is much deeper with a wider range of label quantity, which indicates a higher level of data heterogeneity. Furthermore, to intuitively recognize the diversity among different Dirichlet parameters, we quantify the data distribution through Wasserstein distance inspired by~\cite{jiao2020toward} as shown in Fig.~\ref{dirichlet distribution on cifar100}\subref{cifar100-emd}.

\subsubsection{Baselines}
We compare our proposed algorithm FedAgg with the following approaches. FedAvg \cite{mcmahan2017communication} is introduced as the fundamental framework in the field of federated learning. FedAdam \cite{reddi2021adaptive} allocates an adaptive optimizer for the global server and a mini-batch SGD optimizer for the participating clients respectively, which averages the local gradient for adaptively updating the global model. FedProx \cite{li2020federated} introduces a regularization prox-term during the local model training process which provides robust convergence on heterogeneous systems. FedDyn \cite{durmus2021federated} proposes dynamic variants of the primal-dual regularizer to mitigate local inconsistencies for each client at each global iteration leading to efficient training. FedBABU \cite{oh2022fedbabu} and FedGH \cite{yi2023fedgh} both consider optimizing the head of the network to acquire better model performance.

\subsubsection{Local Training Model Architecture} 
In our numerical experiments, we implement five different local models with various architectures for each real-world dataset. (\romannumeral 1) For the MNIST dataset, we first use a linear model with a fully connected layer of 784 input channels and 10 output channels which can be represented as MNIST-Linear. Besides, we also train a local model for the classification of hand-written digits in MNIST by using a convolutional neural network (CNN) and termed it MNIST-CNN. This CNN has two 5 $\times$ 5 convolution layers, with each layer followed by 2 $\times$ 2 max pooling convolution layers, two fully connected layers with 7 $\times$ 7 $\times$ 64 and 512 units, and a ReLU output layer with 10 units. (\romannumeral 2) For the EMNIST-L dataset, we utilize classic LeNet-5 as a federated learning local training model for the gray-scale hand-written letters classification task. LeNet-5 contains two 5 $\times$ 5 convolution layers, each of which is followed by a 2 $\times$ 2 max pooling layer, three fully connected layers with 5 $\times$ 5 $\times$ 32, 120 and 84 units, and two ReLU output layers with 26 units. (\romannumeral 3) As to CIFAR-10 dataset, another CNN is implemented, referred as CIFAR-10-CNN, which contains three 3 $\times$ 3 convolution layers, with each layer followed by 2 $\times$ 2 max pooling, two dropout layers to avoid overfitting, each of which is followed by a fully connected layer with 4 $\times$ 4 $\times$ 64 and 512 units respectively and a ReLU output layer with 10 units. (\romannumeral 4) We conduct numerical on the CIFAR-100 dataset with standard VGG-11, the same as the experiments of \cite{mcmahan2017communication, kundu2021hire} to achieve well training performance. As the feature dimensions $K^{\prime}$ before classifiers cannot meet the assumptions in FedGH, thus, similar to \cite{zhang2024upload}, we add an average pooling layer before classifiers and default $K^{\prime}=512$.

\begin{table}[t]
\setlength{\abovecaptionskip}{0pt} 
\caption{Hyperparameter of all Datasets}
\renewcommand\arraystretch{1.0}
\begin{center}
\resizebox{0.48\textwidth}{!}{\begin{tabular}{c|c|c|c|c}
\toprule[1pt]
\textbf{Datasets+Local Model}&\textbf{Learning Rate}&\textbf{Batchsize}&\textbf{Global Iteration} &\textbf{Local Epoch}  \\ \cmidrule[0.5pt](l{1pt}r{0pt}){1-5}

\textbf{MNIST+Linear} & 0.01 & 32  & 30 & 3 \\ \cmidrule[0.5pt](l{1pt}r{0pt}){1-5}

\textbf{MNIST+MNIST-CNN} & 0.01 & 32  & 30 & 3 \\ \cmidrule[0.5pt](l{1pt}r{0pt}){1-5}

\textbf{EMNIST-L+LeNet-5} & 0.1 & 64  & 50 & 3 \\ \cmidrule[0.5pt](l{1pt}r{0pt}){1-5}

\textbf{CIFAR-10+CIFAR-10-CNN}& 0.1 & 32  & 100 & 3 \\ \cmidrule[0.5pt](l{1pt}r{0pt}){1-5}

\textbf{CIFAR-100+VGG-11}& 0.01 & 64  & 200 & 3 \\ 
\bottomrule[1pt]
\end{tabular}}
\label{hyperparameter}
\end{center}
\vspace{-12pt}
\end{table}

\subsubsection{Hyperparameters Settings} 
For all datasets, we default local training epoch $L$ = 3 and weight parameter $\alpha$ = 0.1 for Eq.~(\ref{local_model_aggregation}). Specifically, with regard to the MNIST dataset, we set learning rate $\eta$ = 0.01, global training iteration $T$ = 30, and the batch size is 32. For both EMNIST-L and CIFAR-10 datasets, we configure learning rate $\eta$ as 0.1, global training iteration as $T$ = 50 and $T$ = 100, and batch size as 64 and 32, respectively. Additionally, as to the CIFAR-100 dataset, we set learning rate $\eta$ = 0.01 and share the same batch size with EMNIST-L, in the meantime, global iteration $T$ is configured as 500. We summarize all parameter configurations for each real-world dataset in Table~\ref{hyperparameter}. Specifically, for FedAdam, we default $\beta_{1}$ = 0.9, $\beta_{2}$ = 0.99 and $\tau $ = 0.001, while for FedProx and FedDyn, we set $\mu$ = 0.01 and $\alpha$ = 0.01, respectively.

\begin{figure*}[t]
\setlength{\abovecaptionskip}{2pt} 
\centerline{\includegraphics[width=1.0\textwidth, trim=50 70 50 70,clip]{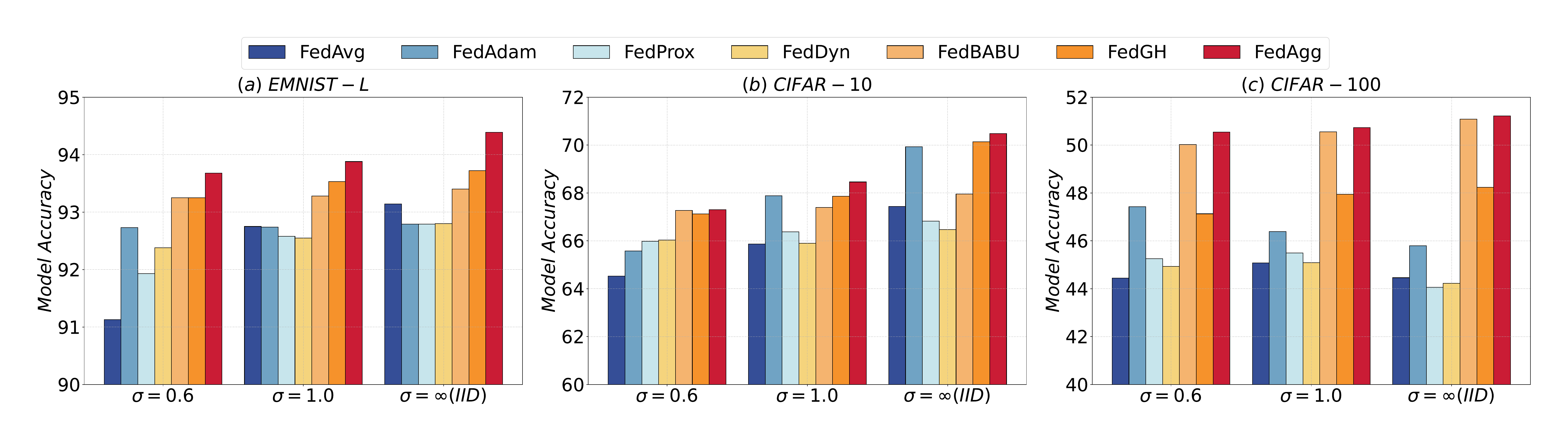}}
\caption{Accuracy on EMNIST-L/CIFAR-10/CIFAR-100 datasets with 100\% of clients across different data heterogeneity.}
\label{EMNIST_Cifar10_Cifar100_barplot}
\vspace{-8pt}
\end{figure*}

\begin{figure*}[t]
\setlength{\abovecaptionskip}{2pt} 
\centerline{\includegraphics[width=1.0\textwidth, trim=50 70 50 70,clip]{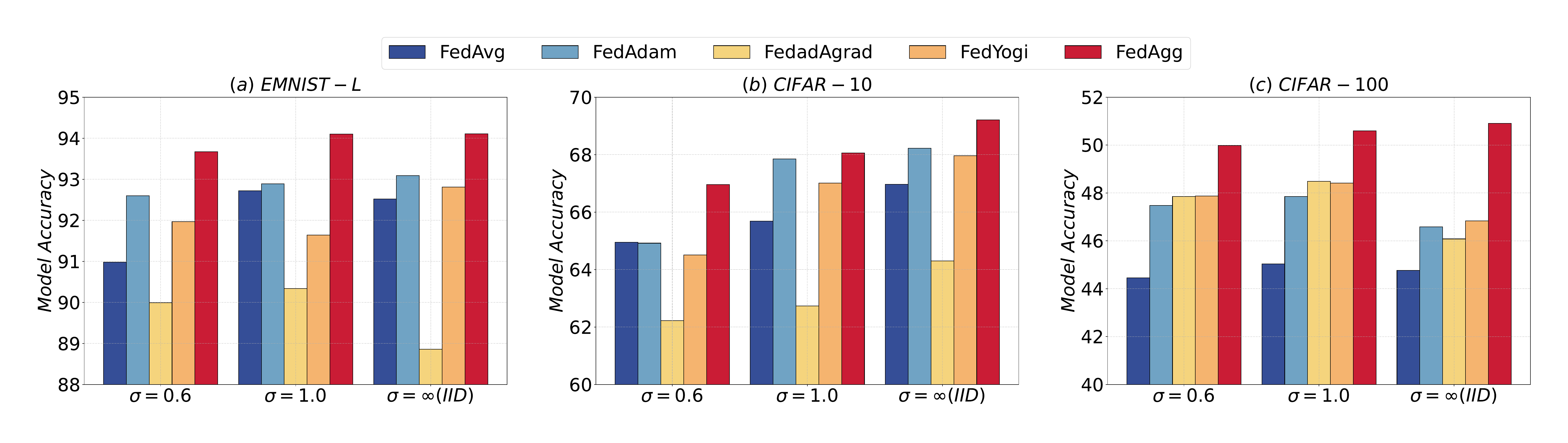}}
\caption{Accuracy on CIFAR-10 dataset with 20\% of participating clients across different adaptive FL algorithms.}
\label{adaptive}
\vspace{-8pt}
\end{figure*}

\subsection{Numerical Results}

\subsubsection{Ablation Study on Number of Participating Clients}
To demonstrate the effectiveness of our proposed algorithm and investigate whether the enhancements introduced by FedAgg remain consistent as the ratio of participating clients increases. Firstly, we partition the four benchmark datasets (i.e., MNIST, EMNIST-L, CIFAR-10, and CIFAR-100) into 100 clients and randomly select 20\% of the total participating clients to participate in the FL training process with dynamically changed local training data and run 30, 50, 100 and 200 global communication iterations, respectively. The main experiment results are displayed in Table~\ref{accuracy}. As shown in Figs. \ref{MNIST_barplot}-\ref{Cifar100_N20}, we visualize the experiment results on all datasets with a 20\% client participating ratio. It is evident that FedAgg dominates other state-of-the-art baseline methods with a faster convergence rate, higher model accuracy, lower training loss, and faster loss descending rate, which demonstrate the immense potential of the adaptive learning rate method in the FL framework. Besides, we conduct experiments with 100\% participating clients to validate the effectiveness of FedAgg under the large-scale federated learning system. According to Table~\ref{accuracy}, in most scenarios, FedAgg dominates the rest baselines and there is a consistent improvement in accuracy with increasing participating ratio. The above numerical experiment results illustrate that our proposed algorithm FedAgg performs well on both small-scale and large-scale FL systems, which demonstrates the capacity of FedAgg for widespread application in real-world federated learning scenarios involving monumental clients and random client participation.

\begin{table}[t]
\setlength{\abovecaptionskip}{3pt} 
\centering
\caption{Comparison to other adaptive FL algorithms.}
\renewcommand\arraystretch{1.2}
\resizebox{0.48\textwidth}{!}{\begin{tabular}
{c|c|ccccc} 
\toprule[1.0pt]
\textbf{Dataset} & \textbf{Distribution} & \textbf{FedAvg} & \textbf{FedAdam} & \textbf{FedAdagrad} & \textbf{FedYogi} & \textbf{FedAgg} \\ \midrule[0.8pt]

\multirow{2}{*}{\multirowcell{2}{\textbf{MNIST} \\ \textbf{+Linear}}}
        & IID & 88.23 & 89.85 & 86.72 & 89.92 & \textbf{90.35} \\ \cmidrule[0.5pt](l{1pt}r{0pt}){2-7}
        
        & Non-IID & 86.28 & 89.08 & 85.81 & 88.57 & \textbf{89.45} \\ \midrule[0.5pt]

\multirow{2}{*}{\multirowcell{2}{\textbf{MNIST} \\ \textbf{+MNIST-CNN}}}
        & IID & 93.59 & 96.55 & 94.32 & 96.48 & \textbf{97.48} \\ \cmidrule[0.5pt](l{1pt}r{0pt}){2-7}
        
        & Non-IID & 86.93 & 87.59 & 86.44 & 89.31 & \textbf{94.91} \\ \midrule[0.8pt]

\multirow{3}{*}{\multirowcell{2}{\textbf{CIFAR-10} \\ \textbf{+CIFAR-10-CNN}}}
        & IID & 66.97 & 68.22 & 64.30 & 67.96 & \textbf{69.21} \\ \cmidrule[0.5pt](l{1pt}r{0pt}){2-7}
        
        & Dir-0.6 & 64.95 & 64.92 & 62.22 & 64.51 & \textbf{66.96} \\ \cmidrule[0.5pt](l{1pt}r{0pt}){2-7}
        
        &  Dir-1.0 & 65.69 & 67.85 & 62.73 & 67.01 & \textbf{68.06} \\ \midrule[0.8pt]

\multirow{3}{*}{\multirowcell{2}{\textbf{EMNIST-L} \\ \textbf{+LeNet-5}}}
        & IID & 92.52 & 93.09 & 88.86 & 92.12 & \textbf{94.11} \\ \cmidrule[0.5pt](l{1pt}r{0pt}){2-7}
        
        & Dir-0.6 & 90.98 & 92.60 & 89.99 & 91.97 & \textbf{93.67} \\ \cmidrule[0.5pt](l{1pt}r{0pt}){2-7}
        
        &  Dir-1.0 & 92.18 & 92.89 & 90.34 & 91.64 & \textbf{94.10} \\ \midrule[0.8pt]

\multirow{3}{*}{\multirowcell{2}{\textbf{CIFAR-100} \\ \textbf{+VGG-11}}}
        & IID & 44.77 & 46.58 & 46.08 & 46.83 & \textbf{50.91} \\ \cmidrule[0.5pt](l{1pt}r{0pt}){2-7}
        
        & Dir-0.6 & 44.45 & 47.48 & 47.85 & 47.87 & \textbf{49.98} \\ \cmidrule[0.5pt](l{1pt}r{0pt}){2-7}
        
        &  Dir-1.0 & 45.04 & 47.85 & 48.49 & 48.41 & \textbf{50.59} \\
\bottomrule[1.0pt]
\end{tabular}}
\label{compared_with_different_adaptive_fl_algorithms}
\vspace{-8pt}
\end{table}

\subsubsection{Ablation Study on Data Heterogeneity}
To demonstrate the robustness of our proposed method across diverse levels of data heterogeneity, we vary the Dirichlet parameter $\sigma \!\in\! \{0.6, \\ 1.0, \infty\}$, where with the descending of $\sigma$, the degree of data heterogeneous increases as shown in Fig.~\ref{dirichlet distribution on cifar100}\subref{cifar100-IID}-\ref{dirichlet distribution on cifar100}\subref{cifar100-dir06}. Besides, $\sigma \!=\! \infty$ means data across clients are IID distribution. Complete experiment results are summarized in Table~\ref{accuracy}. Additionally, we visualize a subset of the numerical results on EMNIST-L, CIFAR-10, and CIFAR-100 datasets with 100\% of participating clients as shown in Fig.~\ref{EMNIST_Cifar10_Cifar100_barplot}. It is evident that our proposed algorithm outperforms the other baseline methods, indicating its robustness and adaptability across diverse data heterogeneity. We also validate our proposed FedAgg algorithm within small-scale FL system scenarios. The results demonstrate that FedAgg still achieves better performance.

\begin{table}[t]
\setlength{\abovecaptionskip}{0pt} 
\caption{Effectiveness on various local model architectures.}
\renewcommand\arraystretch{1.0}
\begin{center}
\resizebox{0.49\textwidth}{!}{\begin{tabular}{c|c|c|c|c|c|c}
\toprule[1pt]
\textbf{Algorithms}&\textbf{LeNet-5}&\textbf{AlexNet}&\textbf{VGG-11} &\textbf{ResNet-18}&\textbf{GoogLeNet}&\textbf{DenseNet121}  \\ \cmidrule[0.5pt](l{1pt}r{0pt}){1-7}

FedAvg & 66.27 & 70.03  & 82.09 & 82.08 & 85.11 & 81.41\\ \cmidrule[0.5pt](l{1pt}r{0pt}){1-7}

\cellcolor{gray!15}FedAgg & \cellcolor{gray!15} 67.20 & \cellcolor{gray!15}70.45  & \cellcolor{gray!15}82.63 & \cellcolor{gray!15}83.15 & \cellcolor{gray!15}86.06 & \cellcolor{gray!15}82.13 \\

\bottomrule[1pt]
\end{tabular}}
\label{different_local_model_architectures}
\end{center}
\vspace{-10pt}
\end{table}

\begin{figure}[t]
\setlength{\abovecaptionskip}{0pt}
\centerline{\includegraphics[width=0.49\textwidth, trim=55 10 50 50,clip]{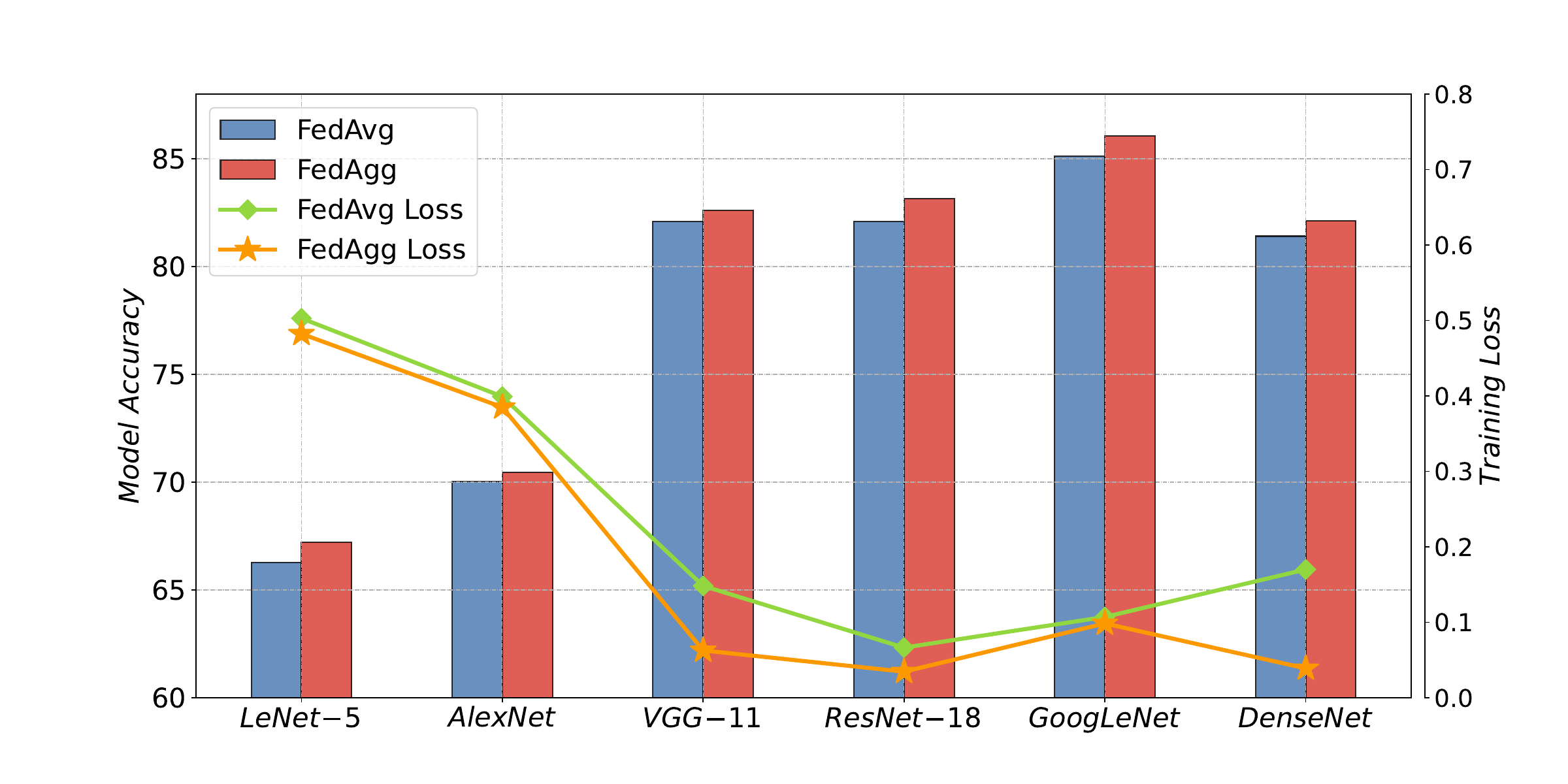}}
\caption{Ablation study of different local model architecture on CIFAR-10 with 20\% of participating clients}
\label{architecture}
\vspace{-8pt}
\end{figure}

\subsubsection{Ablation Study on Extra Adaptive FL Algorithms} \label{extra adaptive baselines}
According to \cite{reddi2021adaptive}, other adaptive algorithms such as FedAdagrad and FedYogi are proposed to improve the model convergence rate under the situation of heterogeneous data. FedAdam employs adaptive learning rates and momentum by leveraging local updates from client devices to efficiently update the global model. FedAdagrad adjusts the learning rate based on the historical gradients of each model parameter, allowing the model to converge faster and achieve better performance. FedYogi, inspired by the Yogi optimizer, incorporates elements of adaptive learning rates and momentum to handle non-convex optimization problems in FL scenarios to improve global model convergence and accuracy. We conduct numerical experiments on CIFAR-10 with 20\% of participating clients. The experiment results are illustrated in Table~\ref{compared_with_different_adaptive_fl_algorithms} and Fig.~\ref{adaptive}. Compared with other adaptive FL algorithms, our proposed FedAgg still performs better with higher accuracy and a faster convergence rate.

\subsubsection{Ablation Study on Different Local Model Architectures} \label{different local model architectures}
We conduct ablation experiments to demonstrate the effectiveness of our proposed algorithm FedAgg across different local model architectures. In addition to the convolutional neural network (CNN) aforementioned, we also implement experiments on LeNet-5, AlexNet, VGG-11, ResNet-18, GoogLeNet, and DenseNet121. Noted that ResNet introduces residual network architecture, GoogLeNet adopts the Inception module and DenseNet121 employs densely connected convolutional networks to effectively alleviate vanishing gradients, enable more efficient feature propagation, and increase the model accuracy. The learning rate for each architecture is set to be 0.1 and performs $T = 100$ iterations of global training on the CIFAR-10 dataset with IID data distribution. Our results are shown in Table~\ref{different_local_model_architectures}. It is worth noting that FedAgg yields consistent enhancements in model performance across various local model architectures and increases the convergence rate of the global model. To observe the improvement of FedAgg across all architectures, we can visualize the intuitional experiment results in Fig.~\ref{architecture}. 

\begin{figure}[t]
\setlength{\abovecaptionskip}{0pt}
\centerline{\includegraphics[width=0.36\textwidth, trim=8 5 5 5,clip]{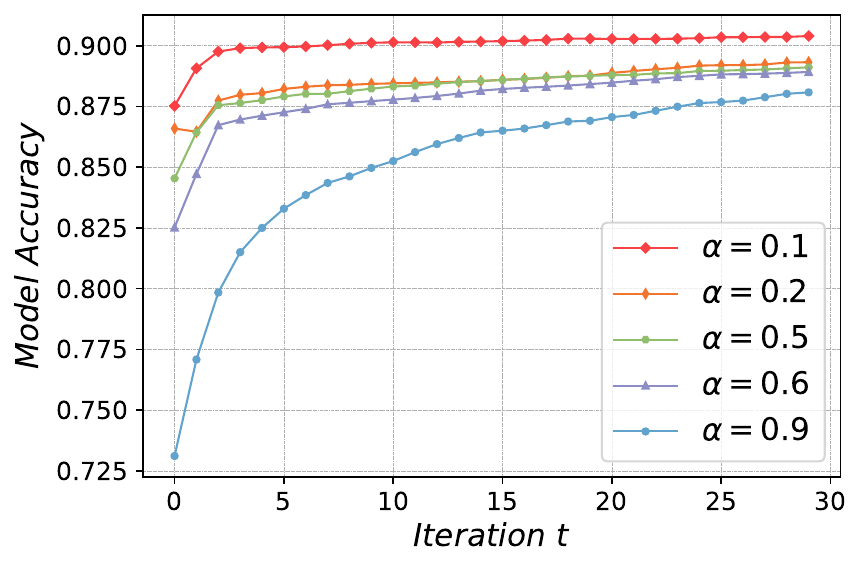}}
\caption{The model performance of our FedAgg algorithm with different aggregation weights $\alpha$ on the MNIST dataset.}
\label{weight}
\vspace{-6pt}
\end{figure}

\subsubsection{Ablation Study on the Aggregation Weight \texorpdfstring {$\alpha$}{}} 
We systematically conduct numerical experiments designed to elucidate the influence exerted by the aggregation weight $\alpha$ in the objective function presented in Eq.~(\ref{objective_mean_field}) on the model efficacy and facilitate the practical application and promotion of FedAgg. As depicted in Fig.~\ref{weight}, the decrement of the hyperparameter $\alpha$ demonstrates that the FL framework accentuates the optimization of the discrepancy between the local model of client $i$ and the average local model, which in turn, bolsters the precision of the global model and expedites the convergence rate. Our findings underscore the significance of meticulous hyperparameter tuning within the FL systems.

\section{Conclusion }\label{Conclusion}
In this paper, we propose an adaptive federated optimization framework with heterogeneous clients for fast model convergence. First, we pioneeringly introduce an aggregated local gradient term into local model update and incorporate a penalty term in each client's objective function to measure its local parameter deviations, thereby alleviating the adverse effect of client heterogeneity. Given the lack of information exchange during local training, we introduce mean-field terms to approximate other clients’ local information over time. Accordingly, we obtain the optimal decentralized adaptive learning rate of each client to enhance the global model training efficiency and expedite convergence. Through rigorous theoretical analysis, we demonstrate the existence and convergence of the mean-field terms. Furthermore, we establish a convergence upper bound for FedAgg, ensuring robust model stability and demonstrating its broad applicability across various scenarios. Finally, extensive experimental results corroborate the superiority of FedAgg compared with other state-of-the-art FL algorithms, underscoring the merits of our proposed adaptive learning rate strategy.



\ifCLASSOPTIONcaptionsoff
  \newpage
\fi



\bibliographystyle{IEEEtran}
\bibliography{references}

\newpage

\appendices

\section{Proof of Proposition~\ref{proposition_lr_linear_combination}} \label{proof_of_proposition_lr_linear_combination}
\begin{proof}
By adopting backward inductive reasoning, we can find the potential rules among $\eta_{i,l}^{t}$, $\boldsymbol{w_{i,l}^{t}}$, $\boldsymbol{\phi_{1,l}^{t}}$ and $\boldsymbol{\phi_{2,l}^{t}}$, $l \!\in\! \{0,  1, \ldots, L\}$. Based on the expression of $\boldsymbol{\lambda(L\!-\!1)}$ and $\boldsymbol{\lambda(L\!-\!2)}$ in Eq.~(\ref{lambda}), we have following derivations. First of all, $\eta_{i,L-1}^{t}$ can be derived from $\boldsymbol{\lambda(L)}$ based on Eq.~(\ref{eta}):
\begin{align} 
\eta_{i,L-1}^{t} &\!=\!  \frac{1}{2 \alpha}( \boldsymbol{\phi_{1,L-1}^{t}} )^\top \boldsymbol{\lambda(L)} \nonumber\\
&\!=\!  \frac{1\!-\!\alpha}{\alpha}(\boldsymbol{\phi_{1,L-1}^{t}})^\top (\boldsymbol{w_{i,L-1}^{t}}\!\!-\!\boldsymbol{\phi_{2,L}^{t}}\!\!-\!\eta_{i,L-1}^{t}\boldsymbol{\phi_{1,L-1}^{t}}) \nonumber  \\
\Rightarrow \eta_{i,L-1}^{t} &\!=\!  \frac{1\!-\!\alpha}{\alpha}\cdot\frac{(\boldsymbol{\phi_{1,L-1}^{t}})^\top(\boldsymbol{w_{i,L-1}^{t}}\!-\!\boldsymbol{\phi_{2,L}^{t}})}{1\!+\!\frac{1\!-\!\alpha}{\alpha}(\boldsymbol{\phi_{1,L-1}^{t}})^\top \boldsymbol{\phi_{1,L-1}^{t}}}. 
\end{align} 

Then, we can figure out the expression of $\eta_{i,L-2}^{t}$ based on $\boldsymbol{\lambda(L\!-\!1)}$ and $\eta_{i,L-1}^{t}$ in the above derivations:
\begin{align} \label{eta_L_minus_2}
\eta_{i,L-2}^{t} &=  \frac{1}{2 \alpha}( \boldsymbol{\phi_{1,L-2}^{t}} )^\top \boldsymbol{\lambda(L \!-\! 1)} \nonumber \\
&= \zeta_{2} (2 \boldsymbol{w_{i,L-1}^{t}} \!-\! \eta_{i,L-1}^{t}\boldsymbol{\phi_{1,L-1}^{t}} \!-\! \zeta_{1}) \nonumber \\
&= \zeta_{2} (2 (\boldsymbol{w_{i,L-2}^{t}}-\eta_{i,L-2}^{t} \boldsymbol{\phi_{1,L-2}^{t}})-\zeta_{1}-\zeta_{3}).\nonumber \\
\Rightarrow \eta_{i,L-2}^{t} &= \frac{\zeta_{2} (2 \boldsymbol{w_{i,L-2}^{t}}-\zeta_{7})}{1+\zeta_{5}+\zeta_{8}}. 
\end{align}
where, 
\begin{align} 
\left\{\begin{array}{l}
\zeta_{1}=\sum_{k=L-1}^{L} \boldsymbol{\phi_{2,k}^{t}}, \zeta_{2}=\frac{1-\alpha}{\alpha}(\boldsymbol{\phi_{1,L-2}^{t}} )^\top, \\[8pt]
\zeta_{3}=\frac{1-\alpha}{\alpha}\frac{( \boldsymbol{\phi_{1,L-1}^{t}})^\top}{1+\zeta_{4}} \zeta_{6}, \zeta_{4}=\frac{1-\alpha}{\alpha}(\boldsymbol{\phi_{1,L-1}^{t}} )^\top \boldsymbol{\phi_{1,L-1}^{t}}, \\[8pt]
\zeta_{5}=\frac{2(1-\alpha)}{\alpha}( \boldsymbol{\phi_{1,L-2}^{t}})^\top \boldsymbol{\phi_{1,L-2}^{t}}, \\[8pt]
\zeta_{6}=(\boldsymbol{w_{i,L-1}^{t}}-\eta_{i,L-2}^{t}\boldsymbol{\phi_{1,L-2}^{t}}-\boldsymbol{\phi_{2,L}^{t}}) \boldsymbol{\phi_{1,L-1}^{t}}, \\[8pt]
\zeta_{7}=\frac{1-\alpha}{\alpha}\frac{(\boldsymbol{\phi_{1,L-1}^{t}})^\top (\boldsymbol{w_{i,L-2}^{t}}-\boldsymbol{\phi_{2,L}^{t}} )\boldsymbol{\phi_{1,L-1}^{t}}+\zeta_{1}}{1+\zeta_{4}}, \\[8pt]
\zeta_{8}=2 (\frac{1-\alpha}{\alpha})^{2}\frac{(\boldsymbol{\phi_{2,L-2}^{t}})^\top(\boldsymbol{\phi_{2,L-1}^{t}})^\top \boldsymbol{\phi_{2,L-2}^{t}}\boldsymbol{\phi_{1,L-1}^{t}}}{1+\zeta_{4}}. \nonumber
\end{array}\right.
\end{align}

Repeating the similar derivation process of deducing $\eta_{i,L-1}^{t}$ and $\eta_{i,L-2}^{t}$, we can acquire the expression of $\eta_{i,l}^{t}$. $\boldsymbol{\phi_{1,l}^{t}}$ and $\boldsymbol{\phi_{2,l}^{t}}$ are both viewed as given functions. Besides, noticed that we can always find a linear function of $\eta_{i,l}^{t}$ expressed by $\boldsymbol{w_{i,l}^{t}}$ and $\eta_{i,q}^{t}$, $q \! \in \! \{l+1, \ldots, L\}$ that we have already calculated before $\eta_{i,l}^{t}$. Therefore, the solution of $\eta_{i,l}^{t}$, $l \!\in\! \{0,1, \ldots, L\!-\!1\}$, can be achieved via an iterative algorithm. \end{proof}

\section{Proof of Proposition \ref{proposition_fixed_point_reach}} \label{proof_proposition_fixed_point_reach}
\begin{proof}
According to the refined local update rule in Eq.~(\ref{local_model_update_with_aggregated_gradient}) and mapping $\Gamma(\{\boldsymbol{w_{i,l}^{t}}, \nabla \! F_{i}(\boldsymbol{w_{i,l}^{t}}) \!\mid\! t \!\in\! \{0,1, \ldots, T\}, l \!\in\! \{0,1, \\ \ldots, L\}\})$ in Eq.~(\ref{mapping}), the local model update can be represented as $(\boldsymbol{w_{i,l+1}^{t}}, \nabla{F_{i}(\boldsymbol{w_{i,l+1}^{t}})}) \!=\! \Gamma_{l}^{t}(\boldsymbol{w_{i,l}^{t}}, \nabla{F_{i}(\boldsymbol{w_{i,l}^{t}})})$. Then, we define the Complete Metric Space $\Omega$ for all model parameter pairs $(\boldsymbol{w}, \nabla{F(\boldsymbol{w}}))$ and the distance function $\Phi$ in the metric space, which measures the difference between two parameter pairs with the $\ell_{2}$-norm, is given by,
\begin{align} \label{distance}
\Phi = \sqrt{\|\boldsymbol{w_{i}}-\boldsymbol{w_{j}}\|_{2}^{2} + \|\nabla{F(\boldsymbol{w_{i}}}) - \nabla{F(\boldsymbol{w_{j}}})\|_{2}^{2}}.
\end{align}

Subsequently, based on Eq.~(\ref{local_model_update_with_aggregated_gradient}) and Eq.~(\ref{distance}), we have the derivations regarding the difference between any parameter pairs $(\boldsymbol{w}, \nabla{F(\boldsymbol{w}}))$ as follows:
\begin{align}
&\Phi(\Gamma_{l}^{t}(\boldsymbol{w_{i,l}^{t}}, \nabla{F_{i}(\boldsymbol{w_{i,l}^{t}})}), \Gamma_{l}^{t}(\boldsymbol{w_{j,l}^{t}}, \nabla{F_{j}(\boldsymbol{w_{j,l}^{t}})})) \nonumber \\ 
=& \sqrt{\|\boldsymbol{w_{i,l+1}^{t}} \!-\! \boldsymbol{w_{j,l+1}^{t}}\|_{2}^{2} \!+\! \|\nabla{F_{i}(\boldsymbol{w_{i,l+1}^{t}})} \!-\! \nabla{F_{j}(\boldsymbol{w_{j,l+1}^{t}})}\|_{2}^{2}} \nonumber \\
\leq&  \sqrt{(\beta^{2} + 1)\|\boldsymbol{w_{i,l+1}^{t}} \!-\! \boldsymbol{w_{j,l+1}^{t}}\|_{2}^{2}} \nonumber \\
\leq& \scriptsize \sqrt{\frac{(\beta^{2} \!+\!1)\delta_{h}}{N}\|(\boldsymbol{w_{i,l}^{t}} \!-\! \boldsymbol{w_{j,l}^{t}}) \!-\!\!\! \sum_{i,j \in N} \!\!\! (\frac{\eta_{i,l}^{t}}{N} \nabla{F_{i}(\boldsymbol{w_{i,l}^{t})}} \!-\!\frac{\eta_{j,l}^{t}}{N} \nabla{F_{j}(\boldsymbol{w_{j,l}^{t})}})\|_{2}^{2}} \nonumber \\
\leq& \footnotesize \sqrt{\frac{(\beta^{2} \!+\!1)\delta_{h}}{N}} \sqrt{\|\boldsymbol{w_{i,l}^{t}} \!-\! \boldsymbol{w_{j,l}^{t}}\|_{2}^{2} \!+\! \|\nabla{F_{i}(\boldsymbol{w_{i,l+1}^{t}})} \!-\! \nabla{F_{j}(\boldsymbol{w_{j,l+1}^{t}})}\|_{2}^{2}} \nonumber \\
\leq& \sqrt{\frac{(\beta^{2} \!+\!1)\delta_{h}}{N}} \Phi((\boldsymbol{w_{i,l}^{t}}, \! \nabla \! F_{i}(\boldsymbol{w_{i,l}^{t}})), (\boldsymbol{w_{j,l}^{t}}, \! \nabla \! F_{j}(\boldsymbol{w_{j,l}^{t}}))), \!\!
\end{align}
where the constant $\sqrt{(\beta^{2} \!+\!1)\delta_{h}/N}$ is selected such that $0 \!\leq\! \sqrt{(\beta^{2} \!+\!1)\delta_{h}/N} \!<\! 1$, thereby ensures that mapping $\Gamma_{l}^{t}$ operates as a contraction mapping. In light of Banach's Fixed Point Theorem, this guarantees the existence of a unique fixed point, denoted as $(\omega_{i,l}^{t*}, \nabla{F_{i}(\omega_{i,l}^{t*})})$, for the iterative computation process. Further, following the inherent contraction property of $\Gamma_{l}^{t}$, which dictates that each subsequent application of $\Gamma_{l}^{t}$ diminishes the distance between consecutive points in the sequence by a factor of $\sqrt{(\beta^{2} \!+\!1)\delta_{h}/N}$. Consequently, for any given positive $\xi$, there exists a positive integer $C$ such that for all $u$, $v \!>\! C$ satisfies: 
\begin{align}
\Phi(\Gamma_{l}^{t}(\boldsymbol{w_{i,u}^{t}}, \nabla{F_{i}(\boldsymbol{w_{i,u}^{t}})}), \Gamma_{l}^{t}(\boldsymbol{w_{i,v}^{t}}, \nabla{F_{i}(\boldsymbol{w_{i,v}^{t}})})) < \xi,
\end{align}
which elucidates that the sequence generated by $\Gamma_{l}^{t}$ is rigorously established as a Cauchy sequence. Drawing upon the foundational tenets of Complete Metric Spaces, it is unambiguously deduced that the sequence converges to the unique fixed point $(\omega_{i,l}^{t*}, \nabla{F_{i}(\omega_{i,l}^{t*})})$. Thus, the theorem holds.~\end{proof}

\section{Proof of Proposition~\ref{proposition_learning_rate_bound}}\label{proof_of_proposition_learning_rate_bound}
\begin{proof} 
Before formal derivation, some prepositive lemma and corollary need to be proved to facilitate the demonstration of Proposition~\ref{proposition_learning_rate_bound}. Then, according to the properties of matrix norm and Definition \ref{definition_mean_field_terms}, we have the following lemmas:

\begin{lemma} \label{lemma 4}
$\|\boldsymbol{\phi_{1,l}^{t}}\|_{2} \leq P$ and $\|\boldsymbol{\phi_{2,l}^{t}}\|_{2} \leq Q$.
\end{lemma}
\begin{proof} From Assumptions \ref{bounded_Gradients}-\ref{assumption_bounded_parameter}, we have:
\begin{align}
\|\boldsymbol{\phi_{1,l}^{t}}\|_{2} &\!=\! \frac{1}{N}\|\sum_{i=1}^{N} \nabla{F_{i} (\boldsymbol{w_{i,l}^{t}})}\|_{2} \nonumber \\
&= \frac{1}{N}\|\nabla{F_{1}(\boldsymbol{w_{1,l}^{t}})} \!+\! \nabla{F_{2} (\boldsymbol{w_{2,l}^{t}})} \!+\! \ldots \!+\! \nabla{F_{N}(\boldsymbol{w_{N,l}^{t}})}\|_{2} \nonumber \\
&\small \stackrel{(a)}{\leq} \!\frac{1}{N} (\|\nabla{F_{1}(\boldsymbol{w_{1,l}^{t}})}\|_{2}
\!+\! \ldots \!+\! \|\nabla{F_{N}(\boldsymbol{w_{N,l}^{t}})}\|_{2}) \!\leq\! P,  \!\!\!\! \\
\|\boldsymbol{\phi_{2,l}^{t}}\|_{2} &\!=\! \frac{1}{N} \|\sum_{i=1}^{N} \boldsymbol{w_{i,l}^{t}}\|_{2} \!=\! \dfrac{1}{N}\|\boldsymbol{w_{1,l}^{t}} \!+\! \boldsymbol{w_{2,l}^{t}} \!+\! \ldots \!+\! \boldsymbol{w_{N,l}^{t}}\|_{2} \nonumber \\
&\stackrel{(a)}{\leq}\! \frac{1}{N} (\|\boldsymbol{w_{1,l}^{t}}\|_{2} \!+\! \|\boldsymbol{w_{2,l}^{t}}\|_{2} \!+\! \ldots \!+\! \|\boldsymbol{w_{N,l}^{t}}\|_{2}) \!\leq\! Q, \!\!
\end{align}
where inequation (a) follows from the triangle inequality.
\end{proof}


Based on Lemma~\ref{lemma 4} and the property of matrix, we can easily get the following inequality: $\|(\boldsymbol{\phi_{1,l}^{t}})^\top\| \leq P$. Similar to the approach of proving Proposition~\ref{proposition_lr_linear_combination}, we adopt Inductive Reasoning to prove Theorem~\ref{proposition_learning_rate_bound} with Lemma~\ref{lemma 4}. Firstly, we prove $\eta_{i, L-1}^{t}$ has an upper bound smaller than 1. Define $\|\boldsymbol{w_{i,l}^{t}}-\boldsymbol{\phi_{2,l}^{t}}\| \leq U $, we have:
\begin{align} \label{eta_L_minus_1_less_1}
\eta_{i,L-1}^{t} &= \frac{1-\alpha}{\alpha}\cdot\frac{(\boldsymbol{\phi_{1,L-1}^{t}})^\top(\boldsymbol{w_{i,L-1}^{t}}-\boldsymbol{\phi_{2,L}^{t}})}{1+\frac{1-\alpha}{\alpha}( \boldsymbol{\phi_{1,L-1}^{t}})^\top \boldsymbol{\phi_{1,L-1}^{t}}} \nonumber \\
&\leq \frac{1-\alpha}{\alpha}\cdot\frac{PU}{1+\frac{1-\alpha}{\alpha}( \boldsymbol{\phi_{1,L-1}^{t}})^\top \boldsymbol{\phi_{1,L-1}^{t}}} \nonumber \\
\Rightarrow \  \eta_{i,L-1}^{t} &\leq \frac{1-\alpha}{\alpha}\cdot\frac{PU}{1+\frac{1-\alpha}{\alpha}P^{2}} = \delta_{i,L-1}^{t} < 1. 
\end{align}

Then, according to Eq.~(\ref{eta_L_minus_2}) and Eq.~(\ref{eta_L_minus_1_less_1}), we have:
\begin{align} \label{eta_L_minus_2_less_1}
\eta_{i,L-2}^{t} &= \frac{1}{2 \alpha}( \boldsymbol{\phi_{1,L-2}^{t}} )^\top \boldsymbol{\lambda(L-1)} \nonumber \\
&= \small \frac{1\!-\!\alpha}{\alpha}(\boldsymbol{\phi_{1,L-2}^{t}} )^\top (2 \boldsymbol{w_{i,L-1}^{t}} \!-\! \eta_{i,L-1}^{t}\boldsymbol{\phi_{1,L-1}^{t}}\!-\!\!  \sum_{k=L-1}^{L} \!\! \boldsymbol{\phi_{2,k}^{t}}) \nonumber \\
&\leq \frac{1-\alpha}{\alpha}P \mid 2Q-\|\eta_{i,L-1}^{t}\boldsymbol{\phi_{1,L-1}^{t}} \!+\! \sum_{k=L-1}^{L} \boldsymbol{\phi_{2,k}^{t}}\| \mid \nonumber \\
\!\!\!\!\!\!\! \Rightarrow \eta_{i,L-2}^{t} &\leq \!\frac{1\!-\!\alpha}{\alpha}P \mid 2Q\!-\!\|\eta_{i,L-1}^{t}\boldsymbol{\phi_{1,L-1}^{t}}\|\!-\! \sum_{k=L-1}^{L} \! \boldsymbol{\phi_{2,k}^{t}}\| \mid \nonumber \\
& \leq \frac{1-\alpha}{\alpha}(\alpha P U - \eta_{i,L-1}^{t} P^{2})^{2} = \delta_{i,L-2}^{t} < 1. 
\end{align}

Thus, based on the process of calculating the upper bound of $\eta_{i, L-1}^{t}$ and $\eta_{i, L-2}^{t}$, we can reasonably induce that all the terms in the expression of adaptive learning rate $\eta_{i,l}^{t}$ (Theorem \ref{theorem_optimal_lr}) is bounded, where $\eta_{i, L-1}^{t}$, $l \!\in\! \{l+1, l+2, \ldots, L-1\}$ can be calculated iteratively similar to $\eta_{i, L-2}^{t}$. In another word, for each $\eta_{i,l}^{t}$, there exist a corresponding upper bound $\delta_{i,l}^{t}$ smaller than 1, where $i \!\in\! \{1, 2, \ldots, N\}$, $l \!\in\! \{0, 1, \ldots, L-1\}$, $t \!\in\! \{0, 1, \ldots, T\}$. 
\end{proof}

\section{Proof of Proposition~\ref{proposition_client_drifting}} \label{proof_of_proposition_client_drifting}
\begin{proof}
Firstly, according to Eq.~(\ref{local_model_update_mean_field}), for any given client $i \in \{0, 1, \ldots, N\}$, we have the following derivations: 
\begin{align}
\|\boldsymbol{w_{i,l+1}^{t}}\!-\!\boldsymbol{\bar w^{t}}\| &= \|\boldsymbol{w_{i,l}^{t}}\!-\!\boldsymbol{\bar w^{t}} \!-\! \eta_{i,l}^{t}  \boldsymbol{\phi_{1,l}^{t}}\| \nonumber \\
&\leq \|\boldsymbol{w_{i,l}^{t}}\!-\!\boldsymbol{\bar w^{t}}\| \!+\! \|\eta_{i,l}^{t}  \boldsymbol{\phi_{1,l}^{t}}\| \nonumber \\
&\leq \|\boldsymbol{w_{i,l}^{t}}\!-\!\boldsymbol{\bar w^{t}}\| \!+\! P \delta_{h}. 
\end{align}

Then, we proceed with a proof by Inductive Reasoning to establish the final result. The induction claim is as follows: $\|\boldsymbol{w_{i,l}^{t}} \!-\! \boldsymbol{\bar w^{t}}\| \!\leq\! l P \delta_{h}$, for any $l \!\in\! \{0, 1, \ldots, L-1\}$. Thus, we demonstrate the basic situation for the local epoch $l = 0$ and $l = 1$. For the local epoch $l = 0$, we have:
\begin{align}
\|\boldsymbol{w_{i,1}^{t}}-\boldsymbol{\bar w^{t}}\| &= \|\boldsymbol{w_{i,0}^{t}}-\boldsymbol{\bar w^{t}} - \eta_{i,0}^{t}  \boldsymbol{\phi_{1,0}^{t}}\| \nonumber \\
&\leq \|\boldsymbol{w_{i,0}^{t}}-\boldsymbol{\bar w^{t}}\| + \|\eta_{i,0}^{t}  \boldsymbol{\phi_{1,0}^{t}}\| \nonumber \\
&\leq \|\boldsymbol{w_{i,0}^{t}}-\boldsymbol{\bar w^{t}}\| + P \delta_{h} \leq P \delta_{h}, 
\end{align}
where $\boldsymbol{w_{i,0}^{t}}=\boldsymbol{\bar w^{t}}$, $i \in \{1, 2, \ldots, N\}$, same as the definition we mentioned in Section \ref{standard_fl_model}. Similarly, for $l = 1$, we have:
\begin{align} 
\|\boldsymbol{w_{i,2}^{t}} \!-\! \boldsymbol{\bar w^{t}}\| &\!=\!  \|\boldsymbol{w_{i,1}^{t}} \!-\! \boldsymbol{\bar w^{t}} \!-\! \eta_{i,1}^{t}  \boldsymbol{\phi_{1,1}^{t}}\| \nonumber \\
&\leq\! \|\boldsymbol{w_{i,1}^{t}} \!-\! \boldsymbol{\bar w^{t}}\| \!+\! \|\eta_{i,1}^{t}  \boldsymbol{\phi_{1,1}^{t}}\| \nonumber \\
&\leq \|\boldsymbol{w_{i,1}^{t}} \!-\! \boldsymbol{\bar w^{t}}\| \!+\! U \delta_{h} \nonumber \\
&\leq \|\boldsymbol{w_{i,0}^{t}} \!-\! \boldsymbol{\bar w^{t}}\| \!+\! 2 P \delta_{h} \!\leq\! 2 P \delta_{h}. 
\end{align}

Similar to the above analysis, for any local epoch $l \geq 2$, the inequality $\|\boldsymbol{w_{i,l}^{t}} - \boldsymbol{\bar w^{t}}\| \leq l P \delta_{h}$ holds. As $l \!\in\! \{0, 1, \ldots, L-1\}$, we have the following derivation:
\begin{align}
\|\boldsymbol{w_{i,l}^{t}}-\boldsymbol{\bar w^{t}}\| \leq l P \delta_{h} \leq L P \delta_{h}. \nonumber
\end{align} 

Hence, the proposition holds. \end{proof}

\section{Proof of Theorem \ref{theorem_model_gap}} \label{proof_of_theorem_model_gap}
\begin{proof}
According to Eq.~(\ref{fresh_local_update}), Eq.~(\ref{local_model_aggregation}) can be rewritten as:
\begin{align} \label{global_model_update}
\boldsymbol{\bar w^{t+1}} &= \sum_{i=1}^{N} \theta_{i} \boldsymbol{w_{i,L}^{t}} =\boldsymbol{w_{i,0}^{t}}-\sum_{i=1}^{N}\sum_{k=0}^{L-1} \theta_{i}\eta_{i,k}^{t} \boldsymbol{\phi_{1,k}^{t}} \nonumber \\
&= \boldsymbol{\bar w^{t}}-\sum_{i=1}^{N}\sum_{k=0}^{L-1} \theta_{i}\eta_{i,k}^{t} \boldsymbol{\phi_{1,k}^{t}}. 
\end{align}

Based on Eq.~(\ref{global_model_update}), the squared $\ell_{2}$-norm of the difference between the global and optimal model parameter is calculated as follows:
\begin{align} \label{parameter_gap}
&\|\boldsymbol{\bar w^{t+1}} - \boldsymbol{w^{*}}\|_{2}^{2} = \|\boldsymbol{\bar w^{t}} -  \boldsymbol{w^{*}}\|_{2}^{2} \nonumber \\
&+ \underbrace{\| \sum_{i=1}^{N}\sum_{k=0}^{L-1} \theta_{i}\eta_{i,k}^{t} \boldsymbol{\phi_{1,k}^{t}} \|_{2}^{2}}_{\Delta_{1}} \underbrace{- 2 \langle \boldsymbol{\bar w^{t}} - \boldsymbol{w^{*}}, \sum_{i=1}^{N}\sum_{k=0}^{L-1} \theta_{i}\eta_{i,k}^{t} \boldsymbol{\phi_{1,k}^{t}} \rangle}_{\Delta_{2}}.
\end{align}
The two terms on the right-hand side of the equality are bounded respectively. For term $\Delta_{1}$, taking the expectation on both sides of the equation, we have the following derivation:
\begin{align}
\mathbb{E}[\Delta_{1}] = \mathbb{E}[\| \sum_{i=1}^{N}\sum_{k=0}^{L-1} \theta_{i}\eta_{i,k}^{t} \boldsymbol{\phi_{1,k}^{t}} \|_{2}^{2}]  
&\stackrel{(a)}{\leq} \delta_{h}^{2} P^{2} \mathbb{E}[\| \sum_{i=1}^{N}\sum_{k=0}^{L-1} \theta_{i} \|_{2}^{2}] \nonumber \\
&\leq \delta_{h}^{2} P^{2} L^{2},
\end{align}
where (a) follows the upper bound of the mean-field term $\boldsymbol{\phi_{1,k}^{t}}$ in Lemma~\ref{lemma 4}. Then, for the term $\Delta_{2}$, we have:
\begin{align}
\mathbb{E}[\Delta_{2}] &= \mathbb{E}[- 2 \langle \boldsymbol{\bar w^{t}} - \boldsymbol{w^{*}}, \sum_{i=1}^{N}\sum_{k=0}^{L-1} \theta_{i}\eta_{i,k}^{t} \boldsymbol{\phi_{1,k}^{t}} \rangle] \nonumber \\ 
&\leq - 2 \delta_{h}  \sum_{i=1}^{N}\sum_{k=0}^{L-1} \theta_{i}\mathbb{E} [\langle \boldsymbol{\bar w^{t}} - \boldsymbol{w^{*}}, \boldsymbol{\phi_{1,k}^{t}} \rangle] \nonumber \\
&\leq - \frac{2 \delta_{h}}{N}  \sum_{i=1}^{N}\sum_{k=0}^{L-1}\sum_{j=1}^{N} \theta_{i}\mathbb{E} [\langle \boldsymbol{\bar w^{t}} - \boldsymbol{w^{*}}, \nabla{F_{j} (\boldsymbol{w_{j,k}^{t}})} \rangle] \nonumber \\
&\stackrel{(a)}{\leq} \!\!- \frac{2 \delta_{h}}{N}\!\! \sum_{i=1}^{N} \! \sum_{k=0}^{L-1} \!\theta_{i}(\mathbb{E}[F_{i}(\boldsymbol{\bar w^{t}}) \!-\! F_{i}(\boldsymbol{w^{*}})] - \frac{1}{2} \beta L^{2}\! P^{2}\! \delta_{h}^{2} N) \nonumber \\
&\leq - \frac{2 \delta_{h} L}{N} \mathbb{E}[F(\boldsymbol{\bar w^{t}}) - F(\boldsymbol{w^{*}})] + \beta P^{2} L^{3} \delta_{h}^{3},
\end{align}
where for (a), the inequality holds based on the following fact:
\begin{align}
&- \sum_{j=1}^{N} \langle \boldsymbol{\bar w^{t}} \!- \boldsymbol{w^{*}}, \nabla{F_{j} (\boldsymbol{w_{j,k}^{t}})} \rangle \nonumber \\ 
= &- \!\!\sum_{j=1}^{N} \langle \boldsymbol{\bar w^{t}} \!-\! \boldsymbol{w_{j,k}^{t}}, \nabla{F_{j} (\boldsymbol{w_{j,k}^{t}})} \rangle \!+\! \langle \boldsymbol{w_{j,k}^{t}} \!-\! \boldsymbol{w^{*}}, \nabla{F_{j} (\boldsymbol{w_{j,k}^{t}})} \rangle  \nonumber \\
\leq& \footnotesize -\!\!\sum_{j=1}^{N} \! F_{i}(\boldsymbol{\bar w^{t}}) \!-\! F_{i}(\boldsymbol{w_{j,k}^{t}}) \!-\! \frac{\beta}{2} \|\boldsymbol{w_{j,k}^{t}} \!-\! \boldsymbol{\bar w^{t}}\|_{2}^{2} \!+\! \langle \boldsymbol{w_{j,k}^{t}} \!-\! \boldsymbol{w^{*}}, \!\!\nabla\! F_{j} (\boldsymbol{w_{j,k}^{t}}) \rangle \nonumber \\
\leq& \small - \sum_{j=1}^{N} \! F_{i}(\boldsymbol{\bar w^{t}}) \!-\! F_{i}(\boldsymbol{w_{j,k}^{t}}) \!+\! F_{i}(\boldsymbol{w_{j,k}^{t}}) \!-\! F_{i}(\boldsymbol{w^{*}}) \!+\! \frac{\beta}{2} \|\boldsymbol{w_{j,k}^{t}} \!-\! \boldsymbol{\bar w^{t}}\|_{2}^{2} \nonumber \\
\leq& - (F(\boldsymbol{\bar w^{t}}) - F(\boldsymbol{w^{*}})) + \sum_{j=1}^{N} \frac{\beta}{2} \|\boldsymbol{w_{j,k}^{t}} - \boldsymbol{\bar w^{t}}\|_{2}^{2} \nonumber \\
\leq& - (F(\boldsymbol{\bar w^{t}}) - F(\boldsymbol{w^{*}})) + \frac{1}{2} \beta L^{2} P^{2} \delta_{h}^{2} N.
\end{align}
Thus, taking the expectation on both sides of Eq.~(\ref{parameter_gap}) and combining the expected upper bound of terms $\mathbb{E}[\Delta_{1}]$ and $\mathbb{E}[\Delta_{2}]$, we obtain the following derivations:
\begin{align} \label{parameter_gap_middle}
\mathbb{E}[\|\boldsymbol{\bar w^{t+1}} \!-\! \boldsymbol{w^{*}}\|_{2}^{2}] \leq&\ \mathbb{E}[\|\boldsymbol{\bar w^{t}}-  \boldsymbol{w^{*}}\|_{2}^{2}] \!+\! \beta P^{2} L^{3} \delta_{h}^{3} \!+\! \delta_{h}^{2} P^{2} L^{2} \nonumber \\
&-\! \frac{2 \delta_{h} L}{N} \mathbb{E}[F(\boldsymbol{\bar w^{t}}) \!-\! F(\boldsymbol{w^{*}})] . \!
\end{align}

Convexity and non-convexity assumptions for the global loss function $F(\boldsymbol{w})$ will lead to different upper bounds. Under the convexity assumption, Eq.~(\ref{parameter_gap_middle}) can be further derived as:
\begin{align} \label{parameter_gap_middle_convexity}
\mathbb{E}[\|\boldsymbol{\bar w^{t+1}} - \boldsymbol{w^{*}}\|_{2}^{2}]  \leq &\ \mathbb{E}[\|\boldsymbol{\bar w^{t}} \!-\! \boldsymbol{w^{*}}\|_{2}^{2}] + \beta P^{2} L^{3} \delta_{h}^{3} \nonumber \\ 
&- \frac{2 \delta_{h} L}{N} \mathbb{E}[F(\boldsymbol{\bar w^{t}}) \!-\! F(\boldsymbol{w^{*}})] + \delta_{h}^{2} P^{2} L^{2} \nonumber \\
\stackrel{(a)}{\leq} & (1 - \frac{\delta_{h} L \psi}{N}) \mathbb{E}[\|\boldsymbol{\bar w^{t}} -  \boldsymbol{w^{*}}\|_{2}^{2}] \nonumber \\
&+ \beta P^{2} L^{3} \delta_{h}^{3} + \delta_{h}^{2} P^{2} L^{2}, 
\end{align}
where (a) follows the Assumption \ref{assumption_convexity}. Through iteratively aggregating both sides of the inequality Eq.~(\ref{parameter_gap_middle_convexity}), we finalize the upper bound of the expected optimality gap after $T$-iteration global training as presented in Eq.~(\ref{model_gap_convex}).

For the non-convexity scenario, we first introduce the following lemma to facilitate our theoretical analysis:
\begin{lemma}\label{lemma_gap}
The local loss function $F_{i}$ satisfies $\beta$-Lipschitz smoothness, for the global model $\boldsymbol{\bar w^{t}}$ at $t$-th iteration, it holds:
\begin{align} \label{gap_iteration}
\mathbb{E}[\|\boldsymbol{\bar w^{t}} - \boldsymbol{w^{*}}\|_{2}] \leq \mathbb{E}[\|\boldsymbol{\bar w^{0}} - \boldsymbol{w^{*}}\|_{2}] + t L P \delta_{h}.
\end{align}
\end{lemma}
\begin{proof}
Firstly, for any global iteration $t \in \{0, 1, \ldots, T\}$, according to Eq.~(\ref{global_model_update}), we have:
\begin{align} \label{gap_middle}
\|\boldsymbol{\bar w^{t+1}} - \boldsymbol{w^{*}}\|_{2} &\leq \|\boldsymbol{\bar w^{t}} - \boldsymbol{w^{*}} - \sum_{i=1}^{N}\sum_{k=0}^{L-1} \theta_{i}\eta_{i,k}^{t} \boldsymbol{\phi_{1,k}^{t}}\|_{2} \nonumber \\
&\leq \|\boldsymbol{\bar w^{t}} - \boldsymbol{w^{*}}\|_{2} + \|\sum_{i=1}^{N}\sum_{k=0}^{L-1} \theta_{i}\eta_{i,k}^{t} \boldsymbol{\phi_{1,k}^{t}}\|_{2} \nonumber \\
&\leq \|\boldsymbol{\bar w^{t}} - \boldsymbol{w^{*}}\|_{2} + L P \delta_{h}.
\end{align}

Through iteratively aggregating both sides of Eq.~(\ref{gap_middle}) with the global iteration $t \!\in\! \{0,1,\ldots, T-1\}$, we can obtain the expression as shown in Eq.~(\ref{gap_iteration}). Hence, the lemma holds. \end{proof} 

Then, according to Lemma~\ref{lemma_gap}, Eq.~(\ref{parameter_gap_middle}) can be further simplified as follows:
\begin{align} \label{parameter_gap_middle_non_convexity}
\mathbb{E}[\|\boldsymbol{\bar w^{t+1}} - \boldsymbol{w^{*}}\|_{2}^{2}]  \leq&\ \mathbb{E}[\|\boldsymbol{\bar w^{t}} \!-  \boldsymbol{w^{*}}\|_{2}^{2}] + \beta P^{2} L^{3} \delta_{h}^{3} \nonumber \\
&- \frac{2 \delta_{h} L}{N} \mathbb{E}[F(\boldsymbol{\bar w^{t}}) - F(\boldsymbol{w^{*}})]  + \delta_{h}^{2} P^{2} L^{2} \nonumber \\
\stackrel{(a)}{\leq}& \mathbb{E}[\|\boldsymbol{\bar w^{t}} - \boldsymbol{w^{*}}\|_{2}^{2}] + \beta P^{2} L^{3} \delta_{h}^{3}  \nonumber \\
&+ \frac{2 \delta_{h} L \|g^\top\|_{2}}{N} \mathbb{E}[\|\boldsymbol{\bar w^{t}} \!-\! \boldsymbol{w^{*}}\|_{2}] + \delta_{h}^{2} P^{2} L^{2} \nonumber \\
\stackrel{(b)}{\leq}& \mathbb{E}[\|\boldsymbol{\bar w^{t}} -  \boldsymbol{w^{*}}\|_{2}^{2}] + \frac{2 t P \delta_{h}^{2} L^{2} \|g^\top\|_{2}}{N}   \nonumber \\
&+ \frac{2 \delta_{h} L \|g^\top\|_{2}}{N}\mathbb{E}[\|\boldsymbol{\bar w^{0}} - \boldsymbol{w^{*}}\|_{2}] \nonumber \\
&+ \beta P^{2} L^{3} \delta_{h}^{3} + \delta_{h}^{2} P^{2} L^{2},
\end{align}
where (a) follows Assumption~\ref{assumption_subgradient} and (b) follows the Lemma~\ref{lemma_gap}. Through iteratively aggregating both sides of the inequality Eq.~(\ref{parameter_gap_middle_non_convexity}), we obtain the upper bound of the expected optimality gap after $T$-iteration global training as shown in Eq.~(\ref{model_gap_nonconvex}).   \end{proof}

\section{Proof of Proposition \ref{proposition_difference}}\label{proof_of_proposition_difference}
To theoretically analyze our proposed algorithm FedAgg, we introduce mathematical Lemmas~\ref{lemma 2}-\ref{lemma 3} \cite{bottou2018optimization, mitra2021achieving} to assist our convergence analysis.
\begin{lemma} \label{lemma 2}
Given any two vectors $ x, y \! \in \! \mathbb{R}^{d}$, the following holds for any $ \gamma \! > \! 0$: 
\begin{align}
\|x+y\|_{2}^{2} \leq(1+\gamma)\|x\|_{2}^{2}+(1+\frac{1}{\gamma})\|y\|_{2}^{2}.
\end{align}
\end{lemma}

\begin{lemma} \label{lemma 3}
Given $m$ vectors $x_{1}, x_{2}, \ldots, x_{m} \in \mathbb{R}^{d}$, according to the Jensen's inequality, we have: 
\begin{align}
\|\sum_{i=1}^{m} x_{i}\|_{2}^{2} \leq m \sum_{i=1}^{m}\|x_{i}\|_{2}^{2}.
\end{align}
\end{lemma}
\begin{proof} 
From the $\beta$-Lipschitz smoothness of $F(\boldsymbol{w})$ in Assumption \ref{assumption_smoothness} and Taylor expansion, we have:
\begin{align} 
& F(\boldsymbol{\bar w^{t+1}}) \!-\! F(\boldsymbol{\bar w^{t}}) \!\leq\!  \langle \boldsymbol{\bar w^{t+1}}\!-\!\boldsymbol{\bar w^{t}}, \nabla{F(\boldsymbol{\bar w^{t}}}) \rangle \!+\!\frac{\beta}{2} \|\boldsymbol{\bar w^{t+1}}\!-\!\boldsymbol{\bar w^{t}} \|_{2}^{2} \nonumber\\
&= \underbrace{\!-\! \sum_{i=1}^{N}\sum_{k=0}^{L-1} \langle\theta_{i}\eta_{i,k}^{t} \boldsymbol{\phi_{1,k}^{t}}, \nabla{F(\boldsymbol{\bar w^{t}})} \rangle}_{\mathcal{T}_{1}} \!+\! \underbrace{ \frac{\beta}{2} \|\sum_{i=1}^{N}\sum_{k=0}^{L-1} \theta_{i}\eta_{i,k}^{t} \boldsymbol{\phi_{1,k}^{t}}\|_{2}^{2}}_{\mathcal{T}_{2}}. 
\end{align}
The two terms on the right-hand side of the inequality are bounded respectively. For the first term $\mathcal{T}_{1}$, we note that: 
\begin{align} \label{T_1}
&\mathcal{T}_{1} = -\sum_{i=1}^{N}\sum_{k=0}^{L-1} \langle\theta_{i}\eta_{i,k}^{t} \boldsymbol{\phi_{1,k}^{t}}, \nabla{F(\boldsymbol{\bar w^{t}})}\rangle \nonumber \\
&=\small - \left\langle \sum_{i=1}^{N}\!\sum_{k=0}^{L-1}\!\! \theta_{i}\eta_{i,k}^{t} (\boldsymbol{\phi_{1,k}^{t}}\!\!-\! \!\nabla{F_{i}(\boldsymbol{\bar w^{t}})})\!+\!\theta_{i}\eta_{i,k}^{t}\! \nabla{F_{i}(\boldsymbol{\bar w^{t}})}, \! \nabla{F(\boldsymbol{\bar w^{t}})}\! \right\rangle \nonumber\\
&\leq \small -\delta_{l} \left\langle  \sum_{i=1}^{N}\!\sum_{k=0}^{L-1}\! \theta_{i} (\boldsymbol{\phi_{1,k}^{t}}\!\!-\!\!\nabla{F_{i}(\boldsymbol{\bar w^{t}})})\!+\!\! \sum_{i=1}^{N}\!\sum_{k=0}^{L-1}\! \theta_{i} \nabla{F_{i}(\boldsymbol{\bar w^{t}})}, \! \nabla{F(\boldsymbol{\bar w^{t}})}\! \right\rangle \nonumber \\
&= \small -\delta_{l} \left\langle \sum_{i=1}^{N}\!\sum_{k=0}^{L-1}\!\theta_{i}(\boldsymbol{\phi_{1,k}^{t}}\!-\!\nabla{F_{i}(\boldsymbol{\bar w^{t}})})\!+\!L\nabla{F(\boldsymbol{\bar w^{t}})}, \nabla{F(\boldsymbol{\bar w^{t}})} \right\rangle \nonumber \\
&=\small  -\delta_{l} \left\langle \sum_{i=1}^{N}\!\sum_{k=0}^{L-1}\! \theta_{i}(\boldsymbol{\phi_{1,k}^{t}}\!\!-\!\!\nabla{F_{i}(\boldsymbol{\bar w^{t}})}),\nabla{F(\boldsymbol{\bar w^{t}})} \right\rangle \!-\! \delta_{l} L \|\nabla{F(\boldsymbol{\bar w^{t}})}\|_{2}^{2} \nonumber\\
&\stackrel{(b)}{\leq} \footnotesize \delta_{l} \left(\|\sum_{i=1}^{N}\! \sum_{k=0}^{L-1}\! \theta_{i}(\boldsymbol{\phi_{1,k}^{t}}\!-\!\nabla{F_{i}(\boldsymbol{\bar w^{t}})})\|_{2} \|\nabla{F(\boldsymbol{\bar w^{t}})}\|_{2} \!-\! L \|\nabla{F(\boldsymbol{\bar w^{t}})}\|_{2}^{2} \right) \nonumber\\
&\stackrel{(a)}{\leq} \footnotesize  \delta_{l}  \left(\sum_{i=1}^{N}\sum_{k=0}^{L-1}\theta_{i}\|\boldsymbol{\phi_{1,k}^{t}}\!-\!\nabla{F_{i}(\boldsymbol{\bar w^{t}})}\|_{2}\|\nabla{F(\boldsymbol{\bar w^{t}})}\|_{2} \!-\! L \|\nabla{F(\boldsymbol{\bar w^{t}})}\|_{2}^{2} \right) \nonumber\\
&\stackrel{(a)}{\leq} \small \frac{\delta_{l}}{N} \left(\sum_{i=1}^{N}\sum_{k=0}^{L-1}\sum_{j=1}^{N}\theta_{i}\|\nabla{F_{j} (\boldsymbol{w_{j,k}^{t}})} \!-\! \nabla{F_{j}(\boldsymbol{\bar w^{t}})}\|_{2} \|\nabla{F(\boldsymbol{\bar w^{t}})}\|_{2} \right) \nonumber \\
&\quad - L \delta_{l} \|\nabla{F(\boldsymbol{\bar w^{t}})}\|_{2}^{2} \nonumber \\
&\stackrel{(c)}{\leq} \small \!\!\frac{\beta\delta_{l}}{N}\!\! \left(\sum_{i=1}^{N}\!\sum_{k=0}^{L-1}\!\sum_{j=1}^{N}\!\theta_{i}\|\boldsymbol{w_{j,k}^{t}}\!\!-\!\boldsymbol{\bar w^{t}}\|_{2}\|\nabla{F(\boldsymbol{\bar w^{t}})}\|_{2} \! \right)  \!-\! L \delta_{l} \|\nabla{F(\boldsymbol{\bar w^{t}})}\|_{2}^{2} \! \nonumber \\
&\stackrel{(d)}{\leq} L^{2} P \beta \delta_{l}\delta_{h} \|\nabla  F( \boldsymbol{\bar w^{t}} )\|_{2}  - L \delta_{l} \|\nabla  F( \boldsymbol{\bar w^{t}} )\|_{2}^{2} \nonumber \\
&\stackrel{(e)}{\leq}  L^{2} P^{2} \beta \delta_{l}\delta_{h} - L \delta_{l} \|\nabla  F( \boldsymbol{\bar w^{t}} )\|_{2}^{2}.
\end{align}
In the above steps, (a) follows from the triangle inequality, (b) follows from the Cauchy-Schwartz inequality, and (c) is a consequence of the $\beta$-Lipschitz smoothness for $F_{i}(\cdot)$ in Assumption \ref{assumption_smoothness}, (d) follows from the bounded client drifting term as proved in Proposition~\ref{proposition_client_drifting}, (e) is the consequence of the bounded gradient in Assumption \ref{bounded_Gradients}. Then, we bound the term $\mathcal{T}_{2}$ as follows:
\begin{align} \label{T_2}
& \mathcal{T}_{2} = \frac{\beta}{2}  \|\sum_{i=1}^{N} \sum_{k=0}^{L-1} \theta_{i}\eta_{i,k}^{t} \boldsymbol{\phi_{1,k}^{t}}\|_{2}^{2} \nonumber \\
&   = \frac{\beta\delta_{h}^{2}}{2}\|\sum_{i=1}^{N}\sum_{k=0}^{L-1} \theta_{i}(\boldsymbol{\phi_{1,k}^{t}} \!-\! \nabla{F_{i}(\boldsymbol{\bar w^{t}})}) \!+\! \sum_{i=1}^{N}\sum_{k=0}^{L-1}\theta_{i}\nabla{F_{i}(\boldsymbol{\bar w^{t}})}\|_{2}^{2} \nonumber \\
&   = \frac{\beta\delta_{h}^{2}}{2}\|\sum_{i=1}^{N}\sum_{k=0}^{L-1} \theta_{i}(\boldsymbol{\phi_{1,k}^{t}} \!-\! \nabla{F_{i}(\boldsymbol{\bar w^{t}})}) \!+\! L\nabla{F(\boldsymbol{\bar w^{t}})} \|_{2}^{2} \nonumber \\
&\stackrel{(a)}{\leq} \beta\delta_{h}^{2} \left(\|\sum_{i=1}^{N}\!\sum_{k=0}^{L-1} \theta_{i}(\boldsymbol{\phi_{1,k}^{t}} \!-\! \nabla{F_{i}(\boldsymbol{\bar w^{t}})})\|_{2}^{2} \! + \! L^{2} \|\nabla{F(\boldsymbol{\bar w^{t}})} \|_{2}^{2} \right) \nonumber \\
&\stackrel{(b)}{\leq} \beta\delta_{h}^{2} \!\left(\!\! N \! \sum_{i=1}^{N} \! \theta_{i}^{2} \|\sum_{k=0}^{L-1}(\boldsymbol{\phi_{1,k}^{t}} \!-\! \nabla{F_{i}(\boldsymbol{\bar w^{t}})})\|_{2}^{2} \!+\! L^{2} \|\nabla{F(\boldsymbol{\bar w^{t}})} \|_{2}^{2} \right) \nonumber \\
&\stackrel{(b)}{\leq} \beta L \delta_{h}^{2} \left( N \sum_{i=1}^{N}\!\sum_{k=0}^{L-1}\! \theta_{i}^{2} \|\boldsymbol{\phi_{1,k}^{t}} \!-\! \nabla{F_{i}(\boldsymbol{\bar w^{t}})}\|_{2}^{2} \!+\! L \|\nabla{F(\boldsymbol{\bar w^{t}})} \|_{2}^{2} \right) \nonumber \\
&\stackrel{(b)}{\leq} \footnotesize \beta L \delta_{h}^{2} \left(\!\sum_{i=1}^{N}\! \sum_{k=0}^{L-1}\!\sum_{j=1}^{N}\! \theta_{i}^{2}\|\nabla{F_{j} (\boldsymbol{w_{j,k}^{t}})} \!-\! \nabla{F_{j}(\boldsymbol{\bar w^{t}})}\|_{2}^{2} \!+\! L \|\nabla{F(\boldsymbol{\bar w^{t}})} \|_{2}^{2} \! \right) \nonumber \\
& \stackrel{(c)}{\leq} \beta L \delta_{h}^{2} \left(\sum_{i=1}^{N}\! \sum_{k=0}^{L-1}\!\sum_{j=1}^{N}\!\theta_{i}^{2} \beta \|\boldsymbol{w_{j,k}^{t}} \!-\! \boldsymbol{\bar w^{t}}\|_{2}^{2} \!+\!  L \|\nabla{F(\boldsymbol{\bar w^{t}})} \|_{2}^{2} \right) \nonumber \\
&  \stackrel{(d)}{\leq} \beta^{2} L^{4} N P^{2} \delta_{h}^{4} + \beta L^{2} \delta_{h}^{2} \|\nabla{F(\boldsymbol{\bar w^{t}})} \|_{2}^{2}.
\end{align}
In the above steps, (a) follows the Lemma~\ref{lemma 2} with $\gamma = 1$, (b) follows the Lemma~\ref{lemma 3}, and (c) is a consequence of the $\beta$-Lipschitz smoothness for $F_{i}(\cdot)$ in Assumption \ref{assumption_smoothness}, (d) follows from the bounded client drifting term as proved in Proposition~\ref{proposition_client_drifting}. Combining with the bound in Eq.~(\ref{T_1}) and Eq.~(\ref{T_2}) immediately leads to the Proposition \ref{proposition_difference}.
\end{proof}

\section{Proof of Theorem \ref{theorem_convergence_rate}} \label{proof_convergence_rate}
\begin{proof}
Based on the convergent upper bound of the global loss function with any two consecutive iterations $t$ and $t+1$ as shown in Proposition \ref{proposition_difference}, we have the following derivation:
\begin{align} \label{difference_PL_inequality}
& F(\boldsymbol{\bar w^{t+1}}) \!-\! F(\boldsymbol{\bar w^{t}}) \nonumber \\
\!\!\!\! \leq  & (\beta L^{2} \delta_{h}^{2} \!-\! L \delta_{l})\|\nabla \! F(\boldsymbol{\bar w^{t}})\|_{2}^{2} \!+\! L^{2} P^{2} \beta \delta_{l}\delta_{h} \!+\! \beta^{2} L^{4} N P^{2} \delta_{h}^{4} \nonumber \\
\!\!\!\! \stackrel{(a)}{\leq} & \footnotesize 2 \mu (\beta L^{2} \delta_{h}^{2} \!-\! L \delta_{l}) (F(\boldsymbol{\bar w^{t}}) \!-\! F(\boldsymbol{w^{*}})) \!+\! L^{2} P^{2} \beta \delta_{l}\delta_{h} \!+\! \beta^{2}\! L^{4}\! N\! P^{2}\! \delta_{h}^{4}, \!\!
\end{align}
where (a) follows the Polyak-\L{}ojasiewicz inequality in Lemma \ref{PL_inequality}. Then, taking expectations and subtracting the optimal global loss function $F(\boldsymbol{w^{*}})$ on both sides of Eq.~(\ref{difference_PL_inequality}), we have:
\begin{align} \label{difference_optimal_PL_inequality}
&F(\boldsymbol{\bar w^{t+1}}) - F(\boldsymbol{w^{*}}) \nonumber \\
\leq&\ F(\boldsymbol{\bar w^{t}}) - F(\boldsymbol{w^{*}}) + 2 \mu (\beta L^{2} \delta_{h}^{2} - L \delta_{l}) (F(\boldsymbol{\bar w^{t}}) - F(\boldsymbol{w^{*}})) \nonumber \\
& + L^{2} P^{2} \beta \delta_{l}\delta_{h} + \beta^{2} L^{4} N P^{2} \delta_{h}^{4} \nonumber \\
\leq&\  (1 + 2 \mu (\beta L^{2} \delta_{h}^{2} - L \delta_{l}))(F(\boldsymbol{\bar w^{t}}) - F(\boldsymbol{w^{*}})) \nonumber \\
&  + L^{2} P^{2} \beta \delta_{l}\delta_{h} + \beta^{2} L^{4} N P^{2} \delta_{h}^{4}.
\end{align}

Through iteratively aggregating Eq.~(\ref{difference_optimal_PL_inequality}) at iteration $t \!\in\! \{0, 1,\ldots, T - 1\}$, we can easily obtain the upper bound of the convergence rate of our proposed FedAgg algorithm as expressed in Eq.~(\ref{convergence_rate}). \end{proof}

\section{Proof of Theorem \ref{theorem_convergence_error}} \label{proof_convergence_error}
\begin{proof}
Based on the convergent upper bound of the global loss function with any two consecutive iterations $t$ and $t+1$ as shown in Proposition \ref{proposition_difference}, Then, taking the total expectation on both sides and rearranging, we have the following derivation:
\begin{align} \label{gradient_l2_norm}
\mathbb{E}[\|\nabla F( \boldsymbol{\bar w^{t}} )\|_{2}^{2} ] \leq&\ \frac{\mathbb{E}[F(\boldsymbol{\bar w^{t}})] - \mathbb{E}[F(\boldsymbol{\bar w^{t+1}})]}{L \delta_{l} -  \beta L^{2} \delta_{h}^{2}} \nonumber \\
&+\! \frac{L P^{2} \beta \delta_{l}\delta_{h} + \beta^{2} L^{3} N P^{2} \delta_{h}^{4}}{\delta_{l} -  \beta L \delta_{h}^{2}}.
\end{align}

By iteratively adding both sides of the inequality in Eq.~(\ref{gradient_l2_norm}) at global iteration $t \!\in\! \{0, 1,\ldots, T - 1\}$, we have:
\begin{align} \label{error}
&\sum_{t=0}^{T-1}\! \mathbb{E}[\|\nabla \! F(\boldsymbol{\bar w^{t}})\|_{2}^{2}] \nonumber \\
\leq&\ \frac{F(\boldsymbol{\bar w^{0}}) \!-\! F(\boldsymbol{\bar w^{T}})}{L \delta_{l} \!-\!  \beta L^{2} \delta_{h}^{2}} \!+\! \frac{T(L P^{2} \! \beta \delta_{l}\delta_{h} \!+\! \beta^{2} \! L^{3} \! N P^{2} \! \delta_{h}^{4})}{\delta_{l} \!-\!  \beta L \delta_{h}^{2}} \nonumber \\
\stackrel{(a)}{\leq}&\ \frac{F(\boldsymbol{\bar w^{0}}) - F(\boldsymbol{w^{*}})}{L \delta_{l} \!-\!  \beta L^{2} \delta_{h}^{2}} \!+\! \frac{T(L P^{2} \! \beta \delta_{l}\delta_{h} \!+\! \beta^{2} \! L^{3} \! N P^{2} \! \delta_{h}^{4})}{\delta_{l} \!-\!  \beta L \delta_{h}^{2}},
\end{align}
where (a) follows the inequality $F(\boldsymbol{w^{*}}) \!>\! F(\boldsymbol{\bar w^{T}})$. 
By rearranging the expression in Eq.~(\ref{error}), we obtain the upper bound of the convergence error in Eq.~(\ref{convergence_error}). \end{proof}

\end{document}